\documentclass[conference]{IEEEtran} 
\ifCLASSOPTIONcompsoc
  \usepackage[nocompress]{cite}
\else
  \usepackage{cite}
\fi
\ifCLASSINFOpdf
\else
\fi
\usepackage[colorlinks=true,
            linkcolor=red,
            urlcolor=blue,
            citecolor=blue]{hyperref}
\usepackage{amsmath,amssymb,amsfonts}
\usepackage{mathtools}
\usepackage{relsize}
\usepackage{graphicx}
\usepackage[scientific-notation=true]{siunitx}
\usepackage[ruled,vlined,lined]{algorithm2e}
\usepackage{cite}
\usepackage{multirow}
\usepackage[noend]{algpseudocode}
\usepackage[utf8]{inputenc}
\usepackage{textcomp}
\usepackage{latexsym}
\usepackage{etoolbox}
\usepackage{subcaption}
\makeatletter
\usepackage[inline]{enumitem}
\def\BState{\State\hskip-\ALG@thistlm}
\makeatother

\AfterEndEnvironment{table}{\vskip-1ex}

\begin{document}

\title{Unsupervised Shot Boundary Detection for Temporal Segmentation of Long Capsule Endoscopy Videos}

% note the % following the last \IEEEmembership and also \thanks - 
% these prevent an unwanted space from occurring between the last author name
% and the end of the author line. i.e., if you had this:
% 
% \author{....lastname \thanks{...} \thanks{...} }
%                     ^------------^------------^----Do not want these spaces!
%
% a space would be appended to the last name and could cause every name on that
% line to be shifted left slightly. This is one of those "LaTeX things". For
% instance, "\textbf{A} \textbf{B}" will typeset as "A B" not "AB". To get
% "AB" then you have to do: "\textbf{A}\textbf{B}"
% \thanks is no different in this regard, so shield the last } of each \thanks
% that ends a line with a % and do not let a space in before the next \thanks.
% Spaces after \IEEEmembership other than the last one are OK (and needed) as
% you are supposed to have spaces between the names. For what it is worth,
% this is a minor point as most people would not even notice if the said evil
% space somehow managed to creep in.

% The paper headers
\markboth{IEEE Transactions on Medical Imaging}%
{Shell \MakeLowercase{\textit{et al.}}: Bare Demo of IEEEtran.cls for IEEE Journals}
% The only time the second header will appear is for the odd numbered pages
% after the title page when using the twoside option.
% 
% *** Note that you probably will NOT want to include the author's ***
% *** name in the headers of peer review papers.                   ***
% You can use \ifCLASSOPTIONpeerreview for conditional compilation here if
% you desire.

% If you want to put a publisher's ID mark on the page you can do it like
% this:
%\IEEEpubid{0000--0000/00\$00.00~\copyright~2015 IEEE}
% Remember, if you use this you must call \IEEEpubidadjcol in the second
% column for its text to clear the IEEEpubid mark.

% use for special paper notices
%\IEEEspecialpapernotice{(Invited Paper)}

% make the title area

\author{\IEEEauthorblockN{Sodiq Adewole\IEEEauthorrefmark{1},
Philip Fernandes \IEEEauthorrefmark{2},
James Jablonski \IEEEauthorrefmark{1}, 
Andrew Copland \IEEEauthorrefmark{2},
Michael Porter \IEEEauthorrefmark{1},\\
Sana Syed \IEEEauthorrefmark{2}, and
Donald Brown\IEEEauthorrefmark{1}\IEEEauthorrefmark{3}}

\IEEEauthorblockA{\IEEEauthorrefmark{1} Department of Systems and Information Engineering,
University of Virginia,
Charlottesville, VA, USA}

\IEEEauthorblockA{\IEEEauthorrefmark{1} Department of Pediatrics, School of Medicine,
University of Virginia,
Charlottesville, VA, USA}

\IEEEauthorblockA{\IEEEauthorrefmark{3} School of Data Science, 
University of Virginia,
Charlottesville, VA, USA}

% \{\href{mailto:soa2wg@virginia.edu}{soa2wg},
% \href{mailto:eg8qe@virginia.edu}{eg8qe},
% \href{mailto:bs2ux@virginia.edu}{bs2ux},
% \href{mailto:meb2fv@virginia.edu}{meb2fv}, 
% \href{mailto:vs3br@virginia.edu}{vs3br},
% \href{mailto:sy8pa@virginia.edu}{sy8pa}, 
% \href{mailto:brown@virginia.edu}{brown}\}@virginia.edu\vspace{-15pt}
}

\maketitle

\begin{abstract}~Physicians use Capsule Endoscopy (CE) as a non-invasive and non-surgical procedure to examine the entire gastrointestinal (GI) tract for diseases and abnormalities. A single CE examination could last between 8 to 11 hours generating up to 80,000 frames which is compiled as a video. Physicians have to review and analyze the entire video to identify abnormalities or diseases before making diagnosis. This review task can be very tedious, time consuming and prone to error. While only as little as a single frame may capture useful content that is relevant to the physicians' final diagnosis, frames covering the small bowel region alone could be as much as 50,000. To minimize physicians' review time and effort, this paper proposes a novel unsupervised and computationally efficient temporal segmentation method to automatically partition long CE videos into a homogeneous and identifiable video segments. However, the search for temporal boundaries in a long video using high dimensional frame-feature matrix is computationally prohibitive and impracticable for real clinical application. Therefore, leveraging both spatial and temporal information in the video, we first extracted high level frame features using a pretrained CNN model and then projected the high-dimensional frame-feature matrix to lower 1-dimensional embedding. Using this 1-dimensional sequence embedding, we applied the Pruned Exact Linear Time (PELT) algorithm to searched for temporal boundaries that indicates the transition points from normal to abnormal frames and vice-versa. The key novelty of this work is in three (3) folds - first, the automated detection of temporal boundaries in long CE video has not been previously considered. Secondly, the reduction in the computational cost of the temporal boundary detection search by using a lower dimensional frame feature embedding; and lastly, the entire temporal segmentation of the CE videos requiring no supervision from medical expert is a new concept. The output of our model can be easily integrated into any CE video summarization model where physicians only need to review a selected sample frame from each video segment. We experimented with multiple real patients' CE videos and our result showed PCA was superior in capturing the transition between pair of normal and abnormal frames in the video. We also bench-marked with expert provided label, and our system achieved an AUC of 66\% on multiple test videos.\vspace{5pt}
\end{abstract}


\section{Introduction}\label{sec:Introduction}
\vspace{-2pt}
With an estimated 70 million Americans affected by different digestive tract diseases each year, physicians use endoscopy as the non-surgical procedure to visualize and examine the stomach, upper small bowel and colon of a person \cite{iddan2000wireless}. Using an endoscope, a flexible tube which carries light by fibreoptic bundles with attached camera, the physician is able to view pictures of the digestive tract on a color TV monitor. Traditionally, three main endoscopy procedures include gastroscopy, small-bowel endoscopy and colonoscopy. During gastroscopy, also known as the upper endoscopy, an endoscope is easily passed through the mouth and throat and into the esophagus, thereby allowing the physician to view the esophagus and stomach \cite{swain2003wireless}. The small bowel endoscopy advances further and allows visibility into the upper part of the small intestine. Colonoscopy involves passing endoscopes into the colon through the rectum to examine the colon. Small bowel endoscopy is especially limited by how far it can advance into the small bowel, thereby limiting the extent of the physicians' examination. All three traditional methods are also limited due to the invasiveness and discomfort that accompanies them. While there has not been a complete replacement for these traditional procedures, especially when a biopsy (removal of tissue) is necessary, Video Capsule Endoscopy (VCE) has innovatively made the endoscopy procedure a lot less invasive and less uncomfortable.

VCE is currently the standard procedure to examine the entire digestive tract without the invasiveness associated with the traditional gastroscopy, small-bowel endoscopy and colonoscopy procedures. While VCE helps ease diagnosis of many digestive tract diseases, a single capsule endoscopy study can last between 8 - 11 hours generating up to 80,000 images of various sections of the digestive tract. In a typical VCE study, up to 50,000 images are obtained for the small bowel region alone, however, it is possible for pathology of interest to be present in as few as one single frame. Notwithstanding, physicians have to review the entire video in order to identify frames capturing diseases or abnormalities. 

Research efforts on automating analysis of VCE videos have been on for more than two decades and many promising methods and techniques have been developed in literature (See section \ref{related_work}). However, many of the proposed techniques focus on identifying specific abnormalities in individual frame independent of other frames in the video. Secondly, Deep Convolutional Neural Network (DCNN) models \cite{Simonyan2014Sep} are currently state-of-the-art models in medical image analysis \cite{chen2017deep, sali2020hierarchical, malik2021ten} and object recognition including various abnormality detection in VCE video frames \cite{gao2020deep}. However, despite their impressive performance on VCE video data, the variety of possible abnormalities in the gastrointestinal (GI) tract coupled with the wide inter-patient variation as well as the sample inefficiency of DCNN models limits their direct applicability towards developing a fully automated system to review and analyze CE videos.

Secondly, the capsule camera used in CE is propelled down the GI tract through peristaltic movement of the intestinal walls and the output videos have unique properties that tend to degenerate the performance of any generic video analysis technique , leading to high misses in diagnosing diseases. For example, poor illumination, food particles causing occlusion and also unstable peristaltic movement of the GI walls results in frequent camera flip sometimes leading to poor quality video output.

Lastly, many open dataset used in traditional video analysis research have already been manually segmented into short video clips with fixed frame counts of fixed time duration \cite{smeaton2010video, geisler2000open}. Therefore many video analysis techniques, especially deep learning based models \cite{shou2016temporal, gao2017video}, are designed to operate mostly on short video clips. Manually segmenting long video into clips have two (2) main problems: 1) The sequence of frames contained in each video clip cannot be guaranteed to be uncorrelated. Manually segmenting long videos, therefore, will not yield a homogeneous and identifiable segment that can lead to optimal summarization output; 2) When a non-homogeneous video segment is to be summarized, there is a chance of selecting a non-key frame as the representative frame, leading to higher miss-rate in any diagnosis.

%%%%%%%%%%%%%%%%%%%%%%%%%%%%%%%%%%%%%%%%%%%%%%%%%%%%%%%%%%%%%%%%%%%%
\section{Related Work}\label{related_work}
\subsection{VCE Video Analysis}
Analysing CE videos encompasses disease or abnormality detection, quantifying severity of identified diseases, localizing identified abnormalities, and decision making on appropriate intervention by the physician. For more than two decades, researchers have proposed different techniques to automate some of these steps by leveraging both classical image analysis and machine learning techniques \cite{rahim2020survey} as well as more recent and advanced deep learning based methods \cite{rasoul2021feature, adewole2020deep, chen2016wireless, adewole2021lesion2vec}. Prior works on VCE fall into three broad categories: 1) detect specific lesion such as bleeding in \cite{sainju2014automated}, polyp  \cite{mamonov2014automated}, ulcer \cite{yuan2015saliency}, and angioectasia \cite{tsuboi2020artificial, pogorelov2018deep}; 2) abnormal or outlier frame detection where frames with abnormalities are consider outliers \cite{gao2020deep, zhao2010abnormality};  and 3) VCE video summarization where representative frames are selected from the entire video \cite{iakovidis2010reduction, emam2015adaptive, mehmood2014video, mohammed2017sparse, ismail2013endoscopy, chen2016wireless} for review by the experts. 

\subsection{Video Temporal Segmentation}
Temporal segmentation is usually the first step when trying to automate analysis of long videos. The goal is to divide the video stream into a set of meaningful segments or \textit{shots}. Each member frame within a segment are correlated and have visual similarity while each segment will exhibit independence characteristic. Vu et al. proposed a coherent three-stage procedure to detect intestinal contractions in \cite{vu2009detection}. The authors utilized changes in intestinal edge structure of the intestinal folds for contraction assessment. The output is contraction-based shots. Mackiewicz et al. in \cite{mackiewicz2008wireless} utilized three dimension LBP operator, color histogram, and motion vector to classify every 10th image of the video. The final classification result was assessed using a 4-state hidden Markov model for topographical segmentation. In \cite{chen2009developing}, two color vectors that were created with hue and saturation components of HSI model were used to represent the entire video. Spectrum analysis was applied to detect sudden changes in the peristalsis pattern. The authors assumed that each organ has a different peristalsis pattern and hence, any change in the pattern may suggest an event in which a gastroenterologist may be interested. Energy and High Frequency Content (HFC) functions are subsequently used to identify such change while two other specialized features aim to enhance the detection of duodenum and cecum. Zhao et al. \cite{zhao2010abnormality} proposed a temporal segmentation approach based on adaptive non-parametric key-point detection model using multi-feature extraction and fusion. The aim of their work was not only to detect key abnormal frames using pairwise distance, but also to augment gastroenterologist's performance by minimizing the miss-rate and thus, improving detection accuracy. None of these prior works considered the computation cost of the temporal segmentation task and given the complexity of CE videos, the time it takes to run a model may render the solution impracticable. The work presented in this paper is motivated by this challenge. In another work aimed at summarizing VCE, \cite{chen2016wireless} proposed to find transition boundaries in the video using pair-wise similarity between the sequence of frames. A threshold parameter is used to determine the boundaries based on the similarity score between frame pairs. Computing pairwise similarity between video frames can be computationally prohibitive and impracticable in real world clinical setting.

\subsection{Boundary Detection}
Detection of boundaries or transition points (TP) on sequence data \cite{sharma2019data} has been considered in solving many sequence segmentation problems across various applications such as medical condition monitoring \cite{malladi2013online}, climate change detection \cite{reeves2007review}, audio activity segmentation and boundary recognition for silence in speech \cite{chowdhury2012bayesian}, speaker segmentation, scene change detection, and human activity analysis \cite{cho2015multiple}. Other areas where detection and localization of distributional changes in sequence data arises include online sequential time series analysis \cite{aminikhanghahi2017survey, tartakovsky2014sequential}. Essentially, Change Point Detection (CPD) involves partitioning a sequence into several homogeneous temporal segments.

Techniques such as probabilistic sequence models including Hidden Markov Models (HMM) \cite{Frade-2007-9831} or the discriminative counterpart such as Conditional Random Fields \cite{Lafferty2001Jun} are well validated. These probabilistic models require a good knowledge of the transition structure between the segments and also require careful pre-training to yield a competitive performance. This may not be practicable for online applications where data are acquired online \cite{adewole2020dialogue, sharma2019data}. Parametric approaches model the distribution before and after the change based on maximum likelihood framework \cite{Chen2012} while non-parametric methods \cite{csorgHo198820} have been mostly limited to uni-variate data. Kernel-based methods \cite{harchaoui2009kernel} use maximum kernel fisher discriminant ratio as a measure of homogeneity between segments and can achieve good results for moderately multidimensional data or in specific situations where the data lie in a low-dimensional manifold. The approach involves a regularized kernel-based test statistic to determine if: 1) there is a change point in the data and thereafter, the location/instant of the change point. However, the method lack robustness when moving to larger dimensions. Particularly, kernel-based methods are not robust with respect to the presence of contaminating noise and to the fact that the changes in the detected points may only affect a subset of the components of the high-dimensional data.

Algorithms such as \textit{Binary Segmentation (BS)} and dynamic programming \cite{Truong2018Jan, Rohrbeck2013}, can identify locations where there are significant changes in the distribution of a sequence of data through recursive search. However, in order to use these techniques, prior knowledge of the number of change point instances in the sequence is required. The algorithms only try to recursively find the location of these points using maximum likelihood estimation. Also, BS search is the most established in literature. The algorithm is an approximate method with an efficient computational cost of $\mathcal{O}(n log n)$, where $n$ is the number of data points. 
\textit{Dynamic Programming (DP) search} is an exact search method, with a computational cost of $\mathcal{O}(Q_{n}^{2})$, where $Q$ is the max number of change points and $n$ is the number of data points \cite{Truong2018Jan}. DP can also be applied using different kernels such as the linear or Gaussian kernels. 
\textit{Window-Based Search} is an approximate search method that computes the discrepancy between two adjacent windows that move along with signal $y$. When the two windows are highly dissimilar, a high discrepancy between the two values occurs, which is indicative of a change point. Upon generating a discrepancy curve, the algorithm locates optimal change point indices in the sequence \cite{Truong2018Jan}. 
\textit{Pruned Exact Linear Time (PELT)} \cite{killick2012optimal} is an unsupervised CPD technique where no prior knowledge of the number of change point is necessary. Rather the model finds the optimal location as well as count of the change points in the series based on a cost function. In temporally segmenting CE videos, no prior knowledge of the number of boundaries is available. Therefore, we considered this technique as most suitable for our task. Other related methods include Segment Neighbourhood (SN) algorithm \cite{auger1989algorithms} and Optimal Partitioning (OP) algorithm \cite{jackson2005algorithm}.

The key to analysis of video structured data is leveraging both spatial (images) and temporal information in the data. While analysis of CE videos has been on for more than two (2) decades, little to no attention has been paid to the temporal relationship between the sequence of frames in the video. In this work, we consider both \textit{spatial} and \textit{temporal} structure of the video to developed a computationally efficient method to temporally segment our long VCE video with the aim of generating multiple shorter, homogeneous and identifiable video segments that are faster and easier to review and analyse. The output of our model could be applied in other domains and also integrated into long video summarization model.

%%%%%%%%%%%%%%%%%%%%%%%%%%%%%%%%%%%%%%%%%%%%%%%%%%%%%%%%%%%%%%%%%%%%%
\subsection{Problem formulation}\label{problem_formulation}
Let $f_{1}, ...,f_{T}$ be unlabelled sequence of frames in a sample CE video $V$. Our hypothesis test consist of;

\textbf{Step 1:} \\
$$\boldsymbol{H}_{0} \rightarrow P_{f_{1}} = .. = P_{f_{k}} = .. = P_{f_{T}}$$
$$\boldsymbol{H}_{A} \rightarrow \exists \hspace{5pt} 1 < k^{*} < T : P_{f_{1}} = .. = P_{f_{k}} \neq P_{f_{k^{*}+1}} = .. = P_{f_{T}}$$

\textbf{Step 2:}
Estimate $k^{*}$ from the sample if $\boldsymbol{H}_{A}$ is true

\begin{figure}%[b]{0.45\textwidth}
    \centering
    \includegraphics[width=0.5\textwidth]{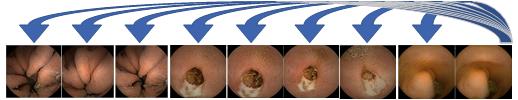}
    \caption{Recursive Search Temporal Shot Boundary in CE Videos}
    \label{fig:recursive_search}
\end{figure}

Figure \ref{fig:recursive_search} shows illustration of the recursive search for a boundary in the contiguous sequence of frames.
%%%%%%%%%%%%%%%%%%%%%%%%%%%%%%%%%%%%%%%%%%%%%%%%%%%%%%%%%%%%%%%%%%%%
Temporal segmentation algorithm:
\begin{algorithm}
%\SetKwData{Data}{Input}
\DontPrintSemicolon
\KwData{VCE video with frames $1:T$; $V = \{f_{1}, f_{2}, ..., f_{T}\}$}
\KwResult{short video segments $\{v_{1}, ..., v_{k}\}$}
\Begin{
 \For{$f_{i}\in V$}{
  Extract Features using CNN: $X_{i} \leftarrow G(f_{i})$\;
  Project each feature vector $\boldsymbol{x_{i}}$ to 1-D embedding $\beta_{i} \leftarrow \boldsymbol{x_{i}}$\; %Compute First $k$ Principal Components (PC): $\boldsymbol{t^{k}_{i}}\leftarrow \boldsymbol{X_{i}}\cdot \boldsymbol{w_{k}}$\;\\
%   Project PC vector $\boldsymbol{t_{k}}$ to 1-D manifold $\beta_{i} \leftarrow \boldsymbol{t_{k_{i}}}$\;
  }
  Concatenate embedding projections $\forall f \in V$; $\boldsymbol{\beta} = \{\beta_{i}, ..., \beta_{T}\}$\;
  Compute transition points $\{\kappa_{i}, ... , \kappa_{m}\}$\;
  Get segments for $V$; \{$v_{j}\}_{j=1}^{k}$\;
  }
\caption{VCE Video Temporal Segmentation algorithm}
\label{sbd_algorithm}
\end{algorithm}
%%%%%%%%%%%%%%%%%%%%%%%%%%%%%%%%%%%%%%%%%%%%%%%%%%%%%%%%%%%%%%%%%%%%%
\section{Methodology}\label{methodology}
\subsection{Overview of Proposed Method}\label{overview}
Algorithm \ref{sbd_algorithm} shows the overview of the proposed technique in this work. Detecting temporal boundaries in long videos allows us to automatically segment long CE videos into short, meaningful, homogeneous and identifiable clips. Our work leverages concept from time series change point analysis \cite{killick2012optimal, Chen2012, harchaoui2009kernel} to detect multiple transition points in a sequence of video frames. CPD methods have been successfully applied on time-series data in one dimension with linear computational time. However, video frame features are usually in higher dimensions, therefore, exponentially increasing the computational cost. In our model, We extracted the frame-features matrix using VGG-19 \cite{Simonyan2014Sep} network model pretrained on large imageNet data and then fine-tune on our VCE video frames. The choice of our architecture is motivated by \cite{adewole2020deep}. Due to the significant class imbalance in the data, we over-sampled the minority classes to minimize the bias of the network towards only the normal class. Thereafter, we projected the frame-features into a 1-dimensional manifold space with the sequence for the entire video appearing like a single time series data. Projecting from p-dimensional video features reduces the computational cost of segmenting the video from $\mathcal{O}(np)$ to $\mathcal{O}(n)$. Thereafter, we applied the Prune Exact Linear Time (PELT) algorithm proposed in \cite{killick2012optimal} to detect multiple transition points in the video. Our model does not require any form of annotation from medical expert. To the best of our knowledge, this is the first work to approach VCE video analysis using concept from CPD model to exploit the temporal information in the sequence of frames. We experimented with multiple embedding methods to compare performance in the segmentation task.

\begin{figure} %[b]{0.45\textwidth}
    \centering
    \includegraphics[width=0.5\textwidth]{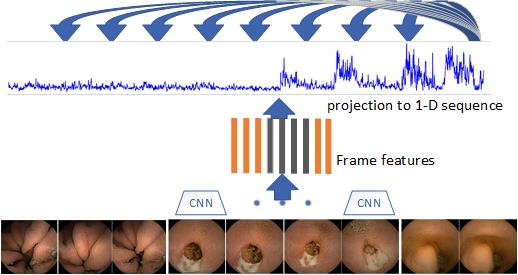}
    \caption{Proposed Temporal Segmentation Pipeline for CE Videos}
    \label{fig:sbd_pipeline}
\end{figure}

\subsection{Lower Dimensional Feature Projection}\label{sec:feature_embedding}
In this section, we describe our approach for embedding the extracted features to a lower 1-dimensional feature. We applied this technique to reduce the computational complexity of finding the temporal boundaries in the video sequence from to $\mathcal{O}(np)$ to a linear time complexity of $\mathcal{O}(n)$. We approached this by projecting the high dimensional frame feature vector to a lower 1-dimensional embedding space. First, we experimented with detecting change boundaries using the high dimensional feature matrix of the video, however, after running for several days on a single video, we recognized the impracticability for real clinical application. Representing abnormalities captured in a VCE image by a single 1-dimensional feature vector is not a trivial task. Therefore, we experimented with other embedding methods to compare performance. Specifically, we experimented with PCA for linear projection and auto-encoder, TSNE, Kernel-PCA with different kernels to account for some non-linearities. We restricted our test to these techniques based on consideration for computational cost and also after experimenting with many manifold learning techniques. We briefly describe each of these embedding techniques below;
%%%%%%%%%%%%%%%%%%
\subsubsection{Principal Component Embedding (PCE)}\label{pce} 
The principal component of a feature matrix extracts the dominant patterns in the matrix in terms of a complementary set of score and loading plots \cite{wold1987principal}. PCA is a linear dimensionality reduction that is used to decompose a multivariate dataset in a set of successive orthogonal components that captures maximum variance in the data. The input data is centered but not scaled for each feature before applying Singular Value Decomposition (SVD). The computational efficiency and speed of PC method makes it a very popular option in the machine learning research community. See figure \ref{fig:embedding} for the visualization of a sample video projected on the 1-dimension that explains most variance using 4096-dimensional feature vector extracted from VGG-19.
%%%%%%%%%%%%%%%%%
\subsubsection{Kernel Principal Component Embedding (KPCE)}\label{kpce} 
\begin{figure}%[h!]
    \centering
    \begin{subfigure}[b]{0.5\textwidth}
        \centering
        \includegraphics[trim=170 0 160 0,clip,scale=0.17]{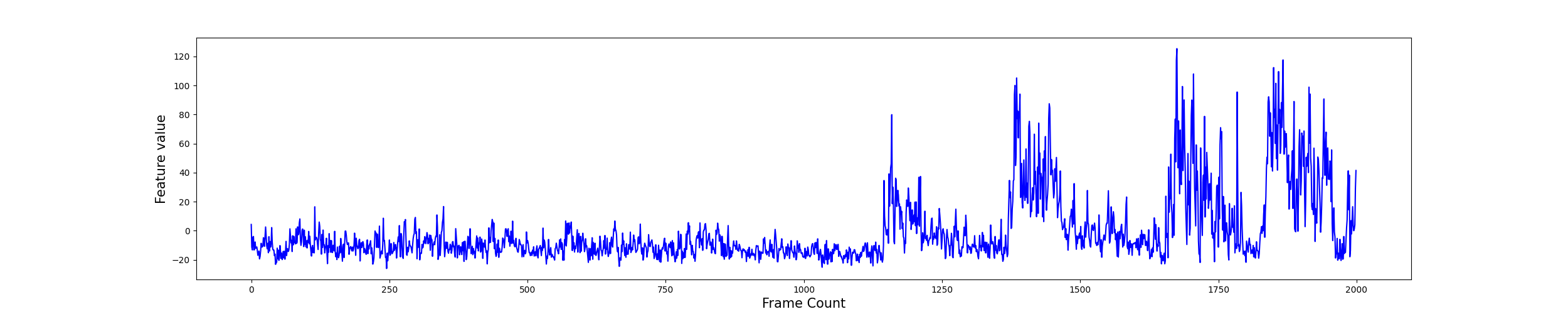}
        \caption{PCA - Linear kernel}
        \label{fig:pca}
    \end{subfigure}
    \begin{subfigure}[b]{0.5\textwidth}
        \centering
        \includegraphics[trim=170 0 160 0,clip,scale=0.17]{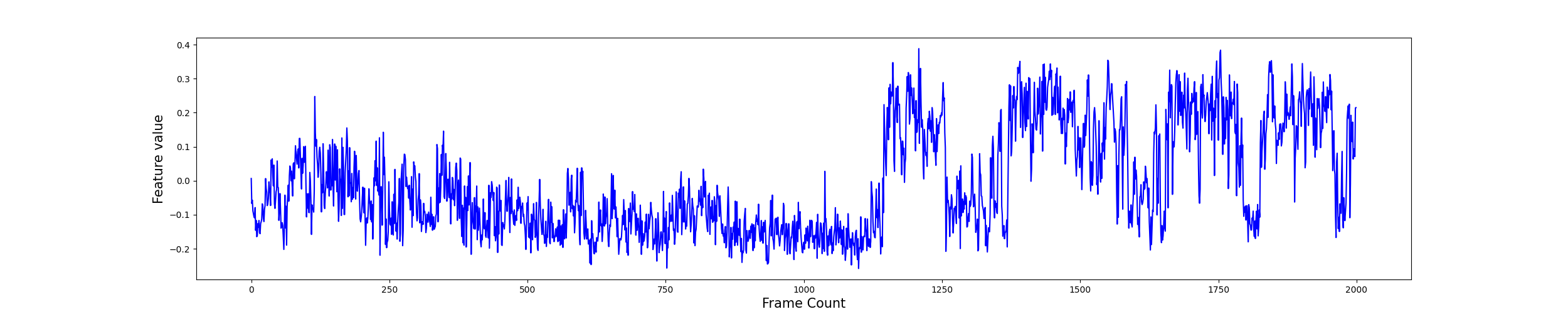}
        \caption{Kernel PCA - Cosine kernel}
        \label{fig:kernel_pca_cos}
    \end{subfigure}
    \begin{subfigure}[b]{0.5\textwidth}
        \centering
        \includegraphics[trim=170 0 160 0,clip,scale=0.17]{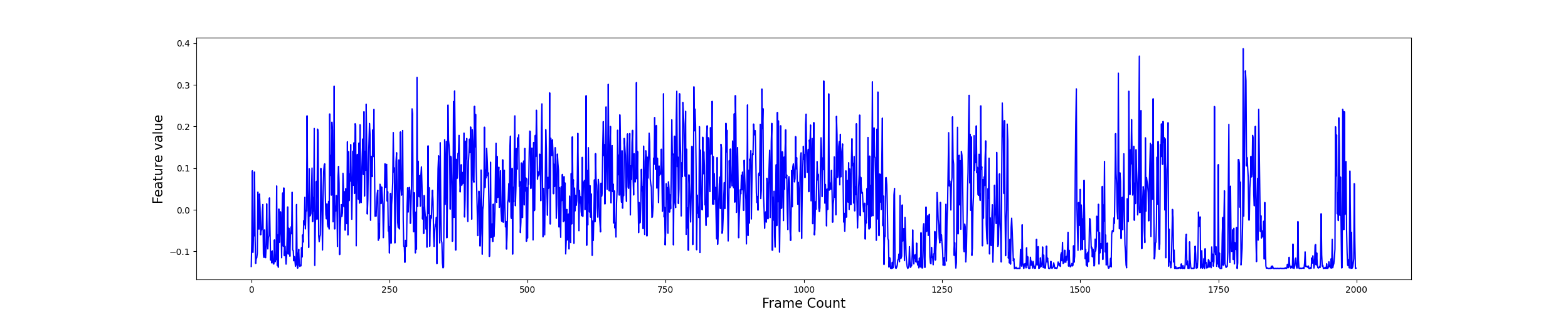}
        \caption{Kernel PCA - Gaussian kernel}
        \label{fig:kernel_pca_rbf}
    \end{subfigure}
    \begin{subfigure}[b]{0.5\textwidth}
        \centering
        \includegraphics[trim=170 0 160 0,clip,scale=0.17]{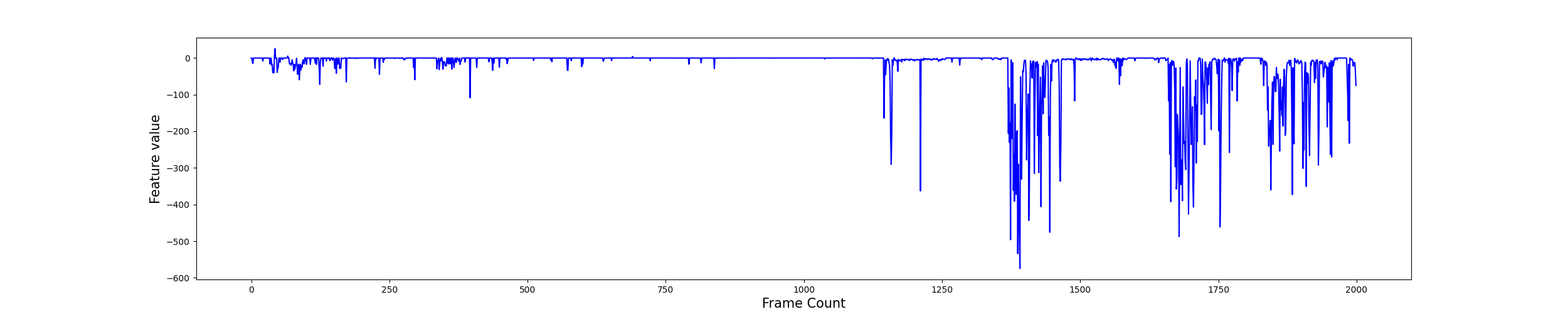}
        \caption{Autoencoder Representation}
        \label{fig:autoencoder}
    \end{subfigure}
    \begin{subfigure}[b]{0.5\textwidth}
        \centering
        \includegraphics[trim=170 0 160 0,clip,scale=0.17]{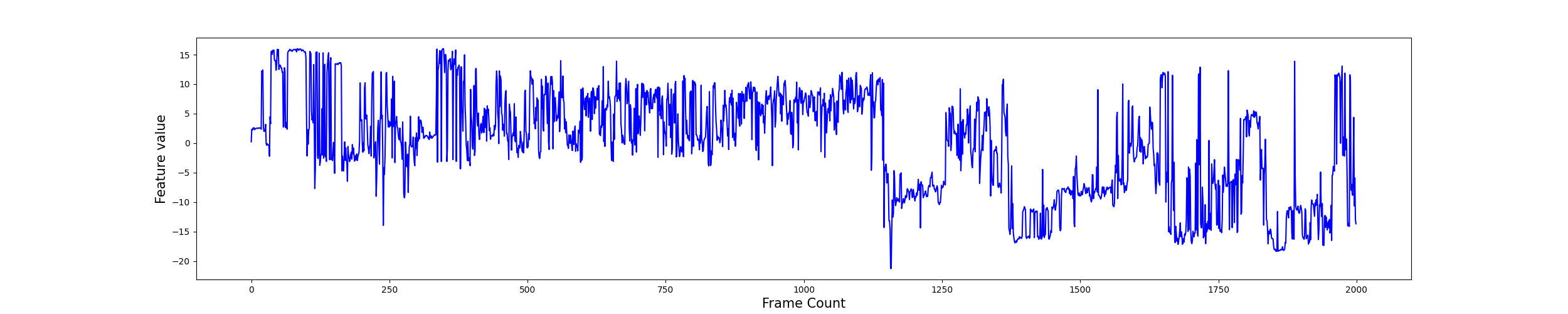}
        \caption{TSNE}
    \end{subfigure}
    \caption{1-D Plot of Sample Video Using Extracted VGG-19 Frame Feature-Matrix}
    \label{fig:tsne}
\label{fig:embedding}
\end{figure}

In order to capture some non-linearities in the embedding, we applied kernel principal component which achieves non-linear dimensionality reduction through the use of kernels. While PCA uses a linear kernel $k(x, y) = X^{T}y$ to construct the eigen-decomposition of the covariance matrix of the data, kernel-PCA uses the kernel trick by mapping the data to a hyperplane with the original linear eigen-decomposition performed in a reproducing kernel hilbert space. We experimented with two (2) different kernels - gaussian and cosine kernels. Figure \ref{fig:kernel_pca_cos} and \ref{fig:kernel_pca_rbf} shows the 1-d projection using the two kernels.

Figures \ref{fig:embedding} shows the visualization of a sample video after projecting to a 1-dimensional embedding space.
The cosine kernel compute the using cosine distance metrics $d(x, y) = \frac{x^{T}y}{||x||\cdot||y||}$. Two objects that are exactly alike have zero distance. The gaussian kernel is an exponential function of the gamma scaled quadratic distance between any two points $k(x, y) = \text{exp}(-\gamma||x - y||^{2}$. The aim of comparing multiple kernels as shown in figure \ref{fig:embedding} is to understand the impact on the sensitivity of the change point algorithm to the structure of the video embedding.
%%%%%%%%%%%%%%%%%%%%%%%%%%%%
\subsubsection{Auto-Encoder} \label{autoencoder}
Auto-encoders learns useful representation without any supervision  \cite{tschannen2018recent}. The goal is to learn a mapping from high-dimensional observations to a lower-dimensional representation space such that the original observations can be reconstructed (approximately) from the lower-dimensional representation. It is a parametric model that is trained using an encoder-decoder neural network architecture. We applied 2-layer architecture and optimized the parameters by minimizing the mean squared loss between the actual frame features and the reconstruction. We used a learning rate of 0.001. The pretrained autoencoder was subsequently used to encode the extracted features for the test videos to a 1-dimensional sequence. While training, we also over-sampled the minority classes to account for the class imbalance as described in \ref{overview}.
%%%%%%%%%%%%%%%%%%%%%%%%
\subsubsection{T-Stochastic Neighborhood Embedding (TSNE)} \label{tsne} 
TSNE \cite{van2008visualizing} uses a probabilistic model to minimize the KL-divergence between the high dimensional input Gaussian distributed feature vector and the lower dimensional t-distributed embedding. We applied TSNE to encode the extracted video features to a 1-dimensional manifold. We set the perplexity parameter to 50 which is similar to the number of nearest neighbour that is used in other manifold learning. Though TSNE can be computationally costly, especially with high dimensional input. We mitigate this problem by first applying PCA on the full video frame features to between 50 - 100 dimension before applying TSNE on the PCA output feature matrix. Figure \ref{fig:tsne} shows the 1-d embedding plot for our test video.
%%%%%%%%%%%%%%%%%%%%%%%%%%%%%%%%%%%%%%%%
\subsection{Shot Boundary Detection (SBD) in CE Video}
In temporally segmenting CE video, we consider temporal boundary as points where there is an occurrence of a pathology between a pair of frames in the sequence. In this paper, we employed the PELT algorithm \ref{pelt} since it requires no supervision in detecting the transition points in the video. The algorithm is derived from the Optimal Programming algorithm but involves a pruning step within the dynamic program to minimize the computational cost. The pruning reduces the computational cost without affecting the exactness of the resulting segmentation making it an ideal candidate for high dimensional video data. The PELT algorithm is able to detect multiple transition points and generally produces quick and consistent results. It solves the penalized detection problem when the number of transition points in the sequence is unknown. By minimizing the log-likelihood cost function in \ref{pelt}, it estimates both the number of transition points as well as location of the change in a sequence of data. The algorithm has a computational cost of $\mathcal{O}(n)$, where $n$ is the number of data points. In our case, $n$ is the number of frames in the video. The PELT algorithm can solve the change point detection problem using different kernels but the most validated is the Gaussian kernel.

On an ordered sequence of frames features $x_{1}, ..., x_{T}$, our SBD model will have $m$ transition points with their positions $\kappa_{1:m} = \{\kappa_{1}, ..., \kappa_{m}\}$; where $1 \leq m \leq T-1$. We specify $\kappa_{0} = 0$ and $\kappa_{m+1} = T$ and assume transition points are ordered such that $\kappa_{i} < \kappa_{j}$. The $m$ transition points will split the data into $m+1$ segments with the $i^{th}$ segment containing $x_{(\kappa_{i-1}+1):\kappa_{i}}$

The algorithm begins by first conditioning on the last point of change, it then iteratively relates the optimal value of the cost function to the cost for the optimal partition of the data prior to the last transition point plus the cost for the segment for the last point to the end of the data \cite{killick2012optimal}. We set $\kappa = \{\kappa:0=\kappa_{0} < \kappa_{1} < \cdot \cdot \cdot < \kappa_{m} < \kappa_{m+1} = T\}$ as the set of possible vectors of transition points for the video. Set $F(0)=-\beta$. The optimal partition is defined as:

\begin{equation}\label{pelt}
\small
\begin{aligned}
    \centering
    F(s) & = \min_{\kappa \in \mathcal{T}_{s}} \sum_{i=1}^{m+1}[C(x_{(\kappa_{i-1}+1):\kappa_{i}})+\beta] \\
    & = \min_{t} \bigg\{\min_{\kappa \in \mathcal{K}_{t}} \sum_{i=1}^{m}[C(x_{(\kappa_{i-1}+1):\kappa_{i}})+\beta] + C(x_{t+1}:n) + \beta \bigg\} \\
    & = \min_{t} \big\{F(t) + C(x_{t+1}:n) + \beta \big\}
\end{aligned}
\end{equation}

Where $C$ is a cost function for the $i^{th}$ segment; $\beta_{m}$ is a regularizer to guard against over fitting which essentially determines how many transition points the algorithm will find. The higher the specified $\beta_{m}$ the less the number of detect transition points forcing the algorithm to minimize the False Positives Rate (FPR). It is important to experiment with this hyper-parameter to make sure increasing the penalty $\beta_{m}$ is not jeopardising the ability to detect true transition points or true positives (TP).

% \begin{equation}
\begin{flalign}\label{cost}
    &C(x_{(\kappa_{i-1}+1):\kappa_{i}}) = \\ \nonumber
    & (\kappa_{i} - \kappa_{i-1}) \left(\text{log}(2\pi) + \text{log} \left(\frac{\sum_{\kappa_{i-1}+1}^{\kappa_{i}}(x_{i} - \mu)^{2}}{\kappa_{i} - \kappa_{i-1}} \right) + 1\right) &&
\end{flalign}
% \end{equation}

$C$ is chosen as twice the negative log-likelihood as in eq. \ref{cost} and the minimum segment length $\kappa_{i-1} - \kappa_{i} \geq 1$.

% \subsection{Temporal Segmentation of VCE Video}
% We assume that a sequence of observed frames features $\{x_{1}, x_{2}, ..., x_{T}\}$ can be divided into non-overlapping, homogeneous segments $\{v_{1}, ... v_{\kappa} \}$ where for each partition $v_{i}$, frame features are correlated and captures the same abnormality.

% Temporally segmenting long videos can be very challenging, the problem is much more complicated in VCE videos where visual transition points does not necessarily indicate pathological event. Hard to find abnormalities such as angioectasia makes detecting transition from no-pathology to a pathological change point. VCE videos have peculiar non-uniform characteristics and inter-frame variations. Such variations may be due to poor lighting in a particular region, food particles causing occlusion,  inter-patients variations as well as instability of the camera motion due peristaltic movement of bowel. Most times, this leads to degraded and poor quality video frames.

\section{Experiments}\label{experiments}
We conducted experiments using eight (8) VCE videos collected during real clinical examination under the supervision of expert gastroenterology. During review and analysis of CE videos, gastroenterologist are mostly interested in the small bowel region which can only be accessed through VCE and not through any of the other upper or lower endoscopy procedures. Detecting pathological change within the small bowel is a much more difficult problem that detecting transition between regions of the GI tract such as esophagus, stomach and colon. For our experiment, we therefore trim the long video to focus only on the small bowel region. Table \ref{tab:sb_data} shows the number of frames per video covering only the small bowel region after removing other regions such as the upper esophagus, stomach and the lower colon.

We extracted the videos from the RapidReader software program and pre-processed each video into frames. The eight (8) videos were collected from different patients during a clinical endoscopy procedure using the SB3 Given Imaging PillCam capsules. The capsules were equipped with 576 x 576 pixel camera. For each complete video, the small bowel transit time corresponds to approximately $3.93\pm1.43$hr \cite{iobagiu2008colon}. In order to isolate the small bowel region, two endoscopy research scientists annotated each video by identifying the region where the image was captured as well as any disease or abnormality found. After the annotation, the number of frames in the videos is summarized in table \ref{tab:sb_data}.

We randomly selected 5 videos for pre-training our feature extraction model and also to perform pre-training of the autoencoder. We reserved the remaining three (3) videos for testing the entire system. Using videos from completely different patients during testing helps minimize any bias and ensures our approach will generalize to any new unseen patient video data.

%% Data
\begin{table}
\centering
\begin{tabular}{c | c | c}
\textbf{Video ID} & \textbf{Training samples} & \textbf{Testing samples}\\
\hline
\centering
Video 1 & 12,303 &  -\\
Video 2 & 13,177 & - \\
Video 3 & 8,452 & - \\
Video 4 & 23,124 & - \\
Video 5 & 32,181 & - \\
Video 6 & - & 8,701 \\
Video 7 & - & 16,909 \\
Video 8 & - & 10,037 \\
\end{tabular}
\caption{Small Bowel Frame Count for Train and Test videos}
\label{tab:sb_data}
\end{table}

\subsection{Implementation} We developed our entire system using the Pytorch framework on NVIDIA GTX2080 machine. We ensured that all our experiment were run on the same configuration for consistency across the compared techniques. Each of the feature extractors were trained for up to 30 epochs using 0.001 as learning rate and Stochastic Gradient Descent optimization. We also trained the autoencoder to embed the frame-features for about 50 epochs. During each of the pre-training, we over-sampled the minority classes based on the inverse of their proportion in the data. This gave a significant boost to the representation capability of the network on the abnormal frames.

\subsection{Evaluation}
We evaluated the performance of this method based on the AUC-ROC as shown in table \ref{tab:roc}. At each time step $t$, the model predicts whether $t$ is a transition point or not. A transition point is defined when the class of frame at $t-1$ is different from the class of frame at $t$. Using the predicted output, we computed the True Positive and False Positive rates and we applied this in computing the ROC. Each transition point is considered to be a pathological event and so we bench-marked against the ground truth label provided by the medical experts. This is, obviously a very challenging problem as both the change point detection algorithm and the feature-embedding models do not have any information on the statistical property that characterizes any of the pathologies in the video.

\section{Results and Discussion}
\begin{figure}
    \centering
    \includegraphics[trim=0 0 0 0,clip,scale=0.091]{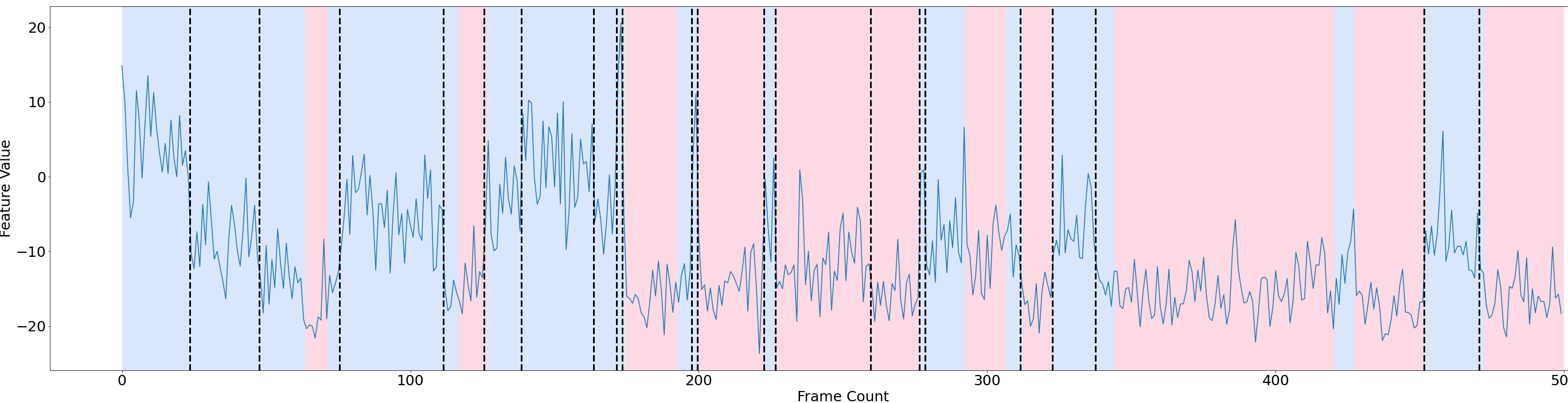}
    \caption{Detected Boundaries vs Ground Truth using PCA @ $\beta = 250$}
    \label{fig:detected_cpd}
\end{figure}

Figure \ref{fig:detected_cpd}, shows experimental results of detected boundaries using PCA embedding and the PELT change point algorithm. Each of the alternating pink-colored intervals are sections of some pathological abnormality. There are points where visually we can observe changes but are not pathological events. These points are due to the camera rotation and flips as it is propelled down the GI tract through peristalsis. This means there is a spatial transition in the content captured by the camera but those changes are not pathological changes.

Experiments on feature extraction also showed that feature extraction capability of the base CNN model is critical to what the boundary detector is able to identify. The representation capability of the base CNN of the diseased-frames will impact the performance of the boundary-detection algorithm. In addition, different CNN architectures showed varying representation performance when applied on different classes of diseases (or lesions). This means, for example Resnet-152 may better represent diffuse bleeding in a frame than VGG-19. Lesions show significant difference both geometrically and in terms of color, texture as well as the surrounding lighting condition. This indicates that the base CNN capabilities are not universal and some architectures better capture some structure than others. 

Table \ref{tab:roc} below shows comparative results using different parametric and non-parametric embedding techniques. Parametric representation frameworks such as auto-encoder are very difficult to train, but are able to capture some non-linearities in the data wherever they successfully train.
% \begin{figure}
%     \centering
%     \includegraphics[trim=280 60 100 60,clip,scale=0.11]{figures/roc_16_250.png}
%     \caption{ROC Plots of Different Embedding @ $\beta=250$}
%     \label{fig:roc}
% \end{figure}

\begin{table}
\centering
\begin{tabular}{l | c }
\textbf{Embedding} & \textbf{AUC-Score} \\
\hline
\centering
PCA & \textbf{0.66} \\
TSNE & 0.49 \\
Kernel-PCA (Cosine) & 0.50 \\
Kernel-PCA (RBF) & 0.50 \\
Auto-Encoder & 0.42 \\
\end{tabular}
\caption{AUC Score of Different Embedding @ $\beta=250$}
\label{tab:roc}
\end{table}

Table \ref{tab:roc} compares the receiver operating characteristics of different embedding techniques with the TPR and FPR aggregated over the test videos. From the table \ref{tab:roc}, while some embedding performed worse than random encoding, PCA with linear kernel achieved an AUC of \textbf{0.66}, outperforming other embedding techniques. Clearly, PCA is able to better encode the frame features to capture more abnormal boundaries than other embedding techniques including the autoencoder (\textbf{0.42}) and TSNE (\textbf{0.49}).

While VGG-19 was able to encode the frames and separate diseased frames from the normal frames, the final pool layer of the model has 4096 features. Embedding this to 1-dimensional vector is not a trivial problem due to the complexity of the CE video frames features and the complex geometry of some abnormalities, such as angioectasia, that may be difficult to detect, even by humans.
%%%%%%%%%%%%%%%%%%%%%%%%%%%%%%%%%%%%%%%%%%%%%%%%%%%%%%%%%%%%%%%%%%%%%%
\subsubsection{Detected Video Boundaries in a Sample Test Video} 
Figure \ref{fig:sbds} below show the detected transition points in the sequence of frames.

\begin{figure}
     \centering
     \begin{subfigure}[b]{0.1\textwidth}
         \centering
         \includegraphics[trim=32 32 32 32,clip,scale=0.04]{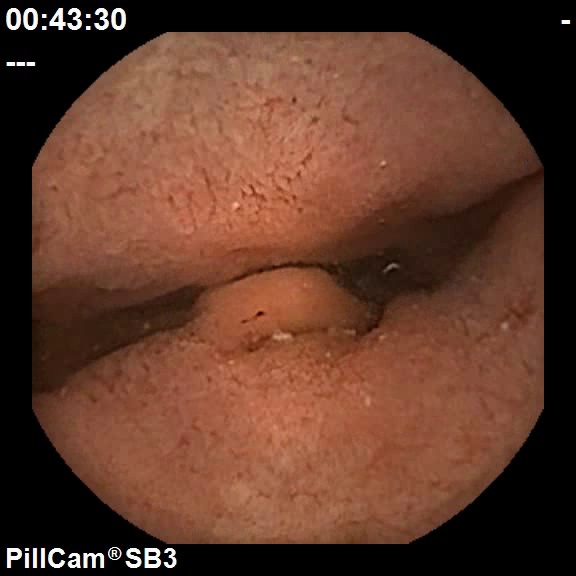}
     \end{subfigure}
     \hspace{-1.35cm}
      \begin{subfigure}[b]{0.1\textwidth}
         \centering
         \includegraphics[trim=32 32 32 32,clip,scale=0.04]{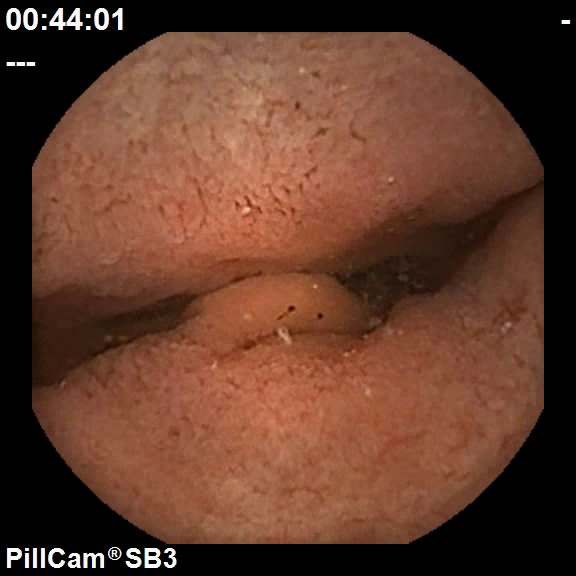}
     \end{subfigure}
     \hspace{-1.35cm}
      \begin{subfigure}[b]{0.1\textwidth}
         \centering
         \includegraphics[trim=32 32 32 32,clip,scale=0.04]{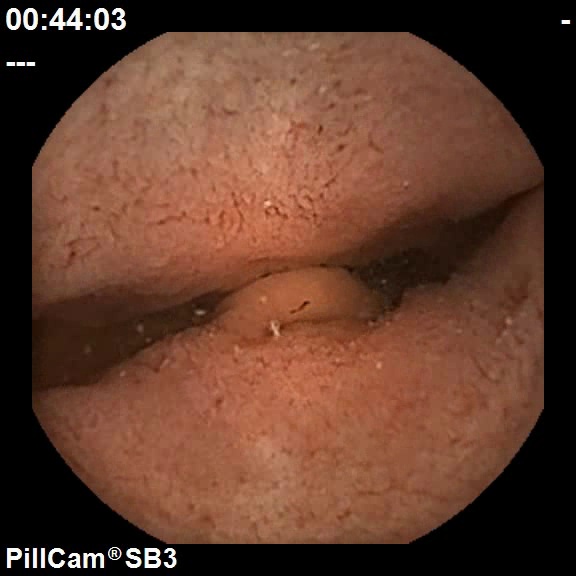}
     \end{subfigure}
    \hspace{-1.35cm}
      \begin{subfigure}[b]{0.1\textwidth}
         \centering
         \includegraphics[trim=32 32 32 32,clip,scale=0.04]{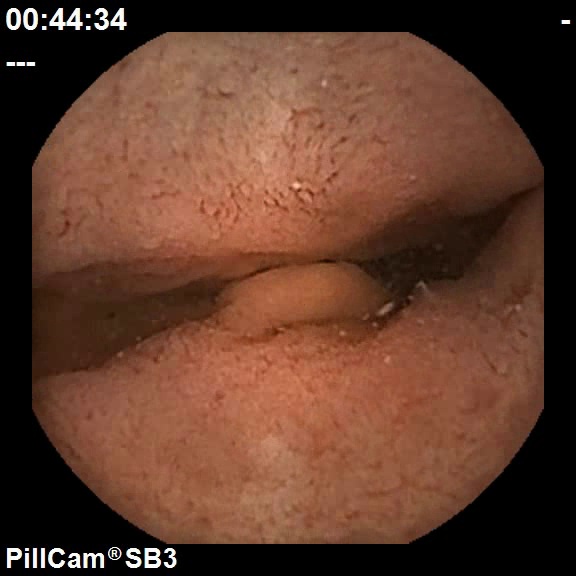}
     \end{subfigure}
    \hspace{-1.35cm}
      \begin{subfigure}[b]{0.1\textwidth}
         \centering
         \includegraphics[trim=32 32 32 32,clip,scale=0.04]{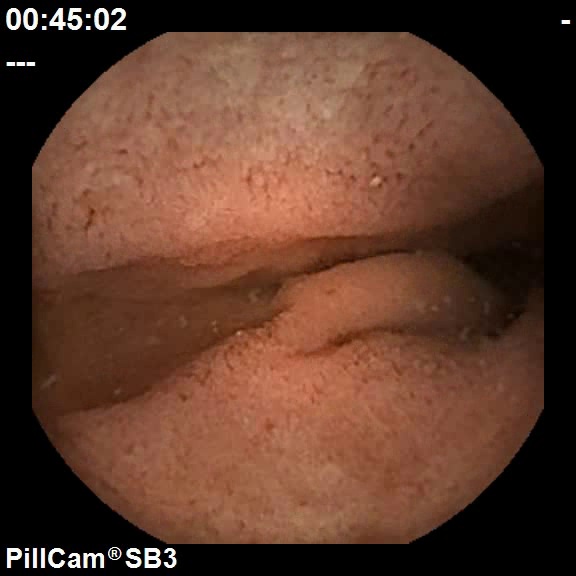}
     \end{subfigure}
    \hspace{-.750cm}
   \rule{1pt}{23pt}
    \hspace{-.750cm}
      \begin{subfigure}[b]{0.1\textwidth}
         \centering
         \includegraphics[trim=32 32 32 32,clip,scale=0.04]{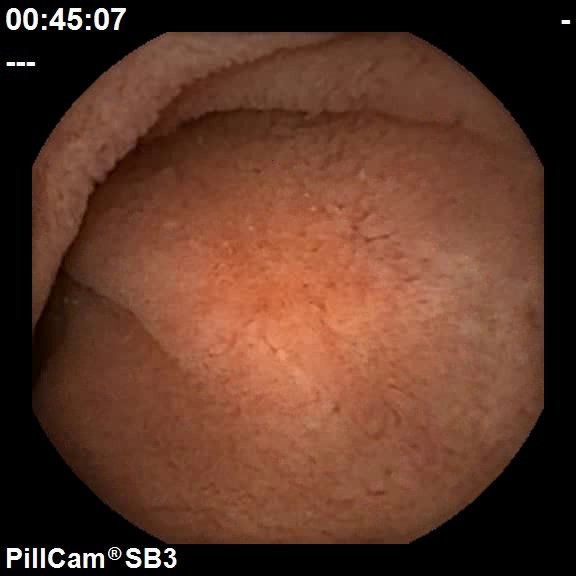}
     \end{subfigure}
    \hspace{-1.35cm}
      \begin{subfigure}[b]{0.1\textwidth}
         \centering
         \includegraphics[trim=32 32 32 32,clip,scale=0.04]{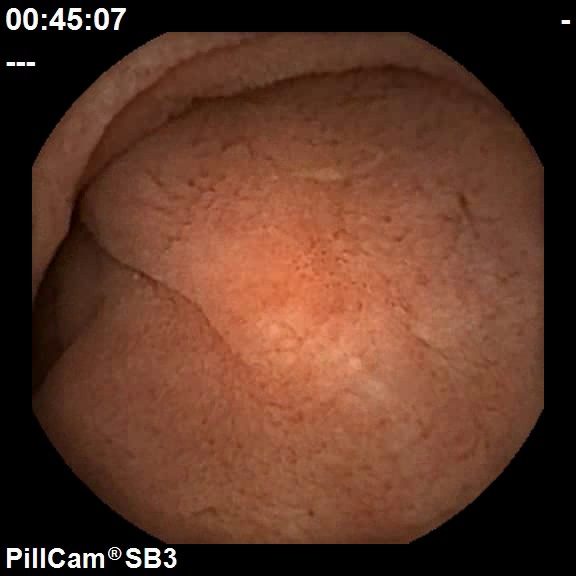}
     \end{subfigure}
    \hspace{-1.35cm}
      \begin{subfigure}[b]{0.1\textwidth}
         \centering
         \includegraphics[trim=32 32 32 32,clip,scale=0.04]{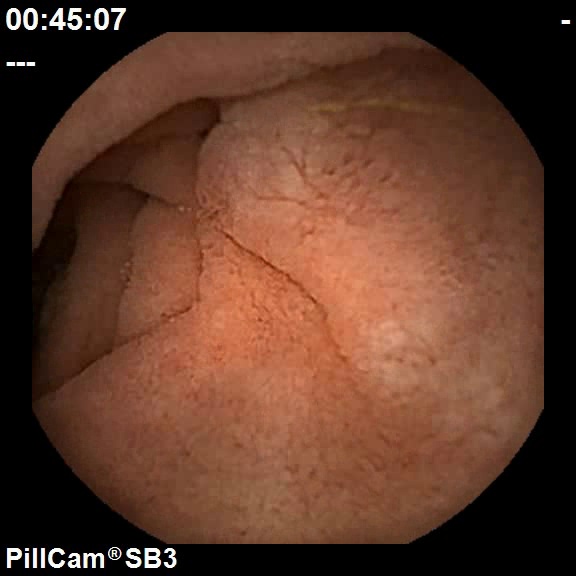}
     \end{subfigure}
    \hspace{-1.35cm}
      \begin{subfigure}[b]{0.1\textwidth}
         \centering
         \includegraphics[trim=32 32 32 32,clip,scale=0.04]{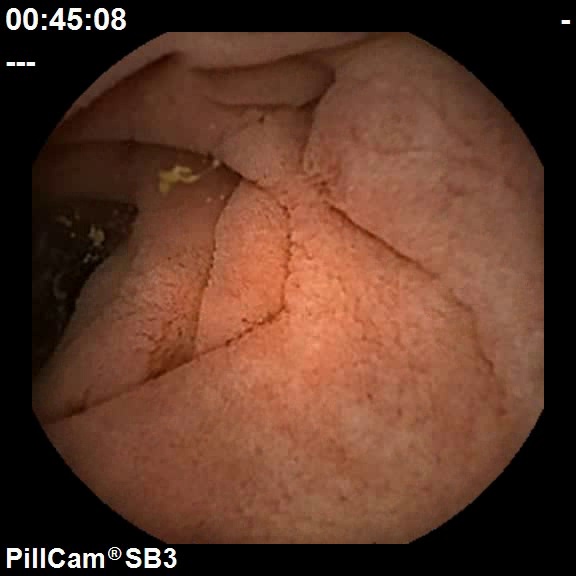}
     \end{subfigure}
    \hspace{-.750cm}
   \rule{1pt}{23pt}
    \hspace{-.750cm}
      \begin{subfigure}[b]{0.1\textwidth}
         \centering
         \includegraphics[trim=32 32 32 32,clip,scale=0.04]{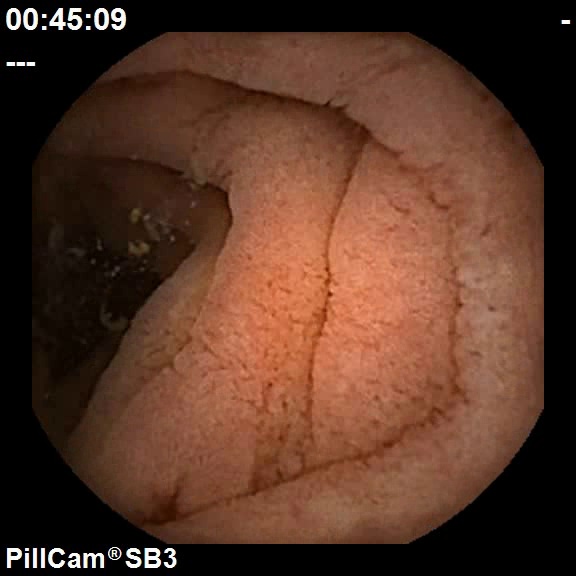}
     \end{subfigure}
     \hspace{-1.35cm}
      \begin{subfigure}[b]{0.1\textwidth}
         \centering
         \includegraphics[trim=32 32 32 32,clip,scale=0.04]{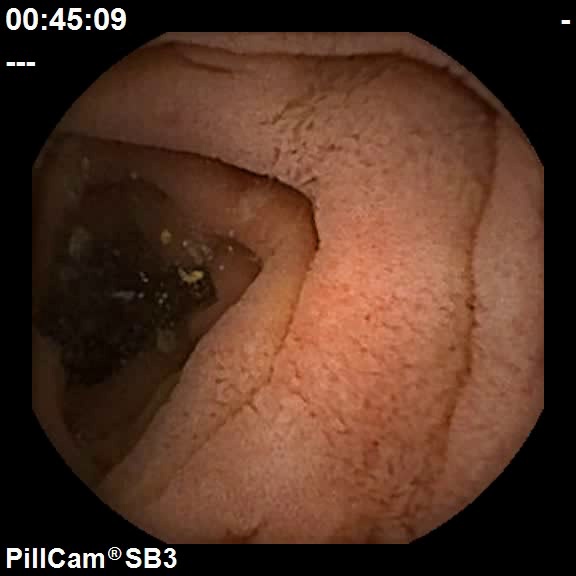}
     \end{subfigure}
     \hspace{-1.35cm}
      \begin{subfigure}[b]{0.1\textwidth}
         \centering
         \includegraphics[trim=32 32 32 32,clip,scale=0.04]{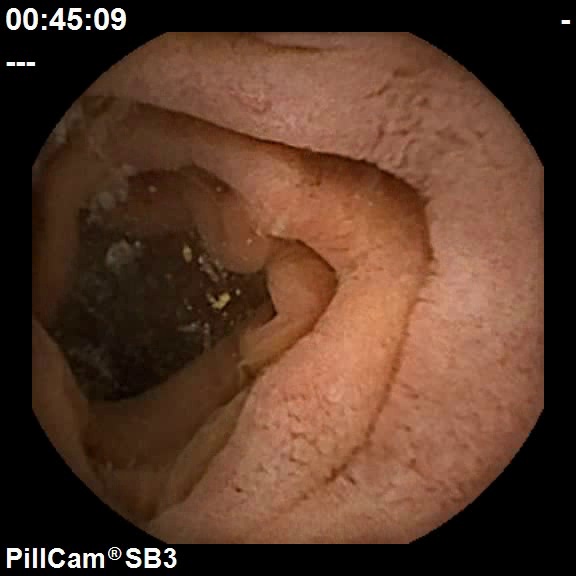}
     \end{subfigure}
    \hspace{-1.35cm}
      \begin{subfigure}[b]{0.1\textwidth}
         \centering
         \includegraphics[trim=32 32 32 32,clip,scale=0.04]{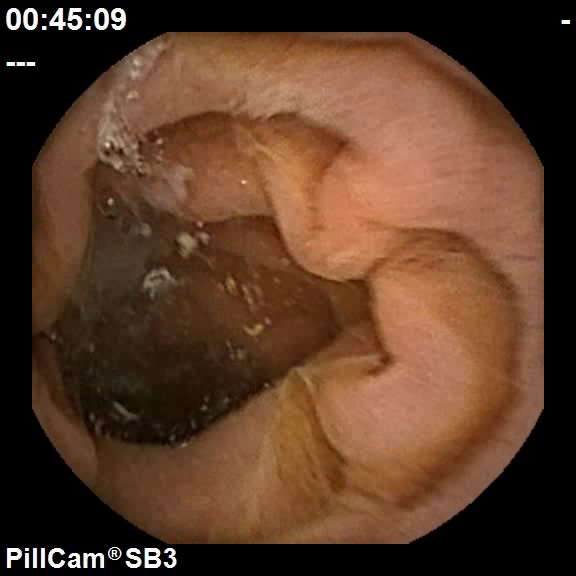}
     \end{subfigure}
    \hspace{-1.35cm}
      \begin{subfigure}[b]{0.1\textwidth}
         \centering
         \includegraphics[trim=32 32 32 32,clip,scale=0.04]{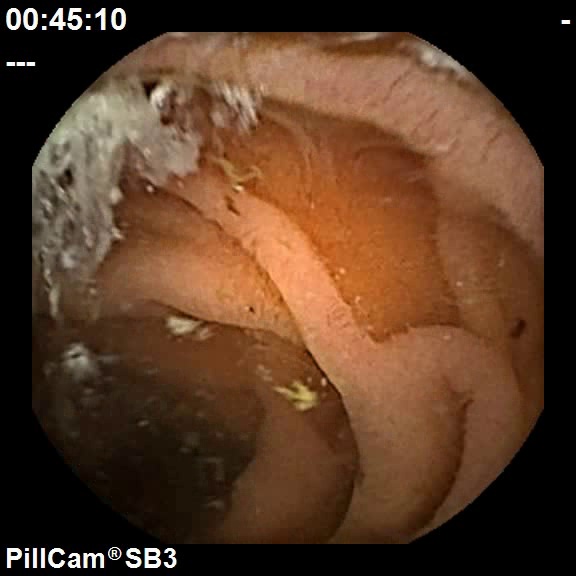}
     \end{subfigure}
    \hspace{-1.35cm}
      \begin{subfigure}[b]{0.1\textwidth}
         \centering
         \includegraphics[trim=32 32 32 32,clip,scale=0.04]{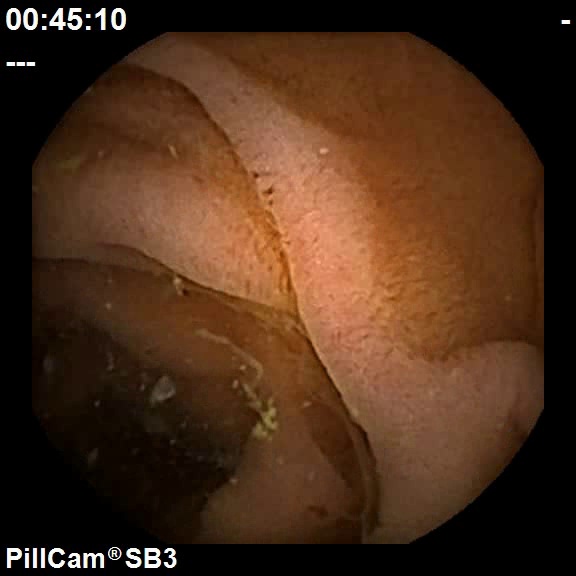}
     \end{subfigure}
    \hspace{-1.35cm}
      \begin{subfigure}[b]{0.1\textwidth}
         \centering
         \includegraphics[trim=32 32 32 32,clip,scale=0.04]{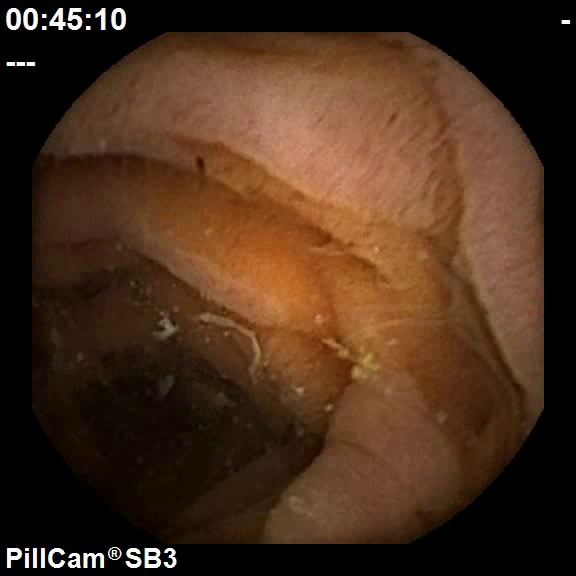}
     \end{subfigure}
    \hspace{-1.35cm}
      \begin{subfigure}[b]{0.1\textwidth}
         \centering
         \includegraphics[trim=32 32 32 32,clip,scale=0.04]{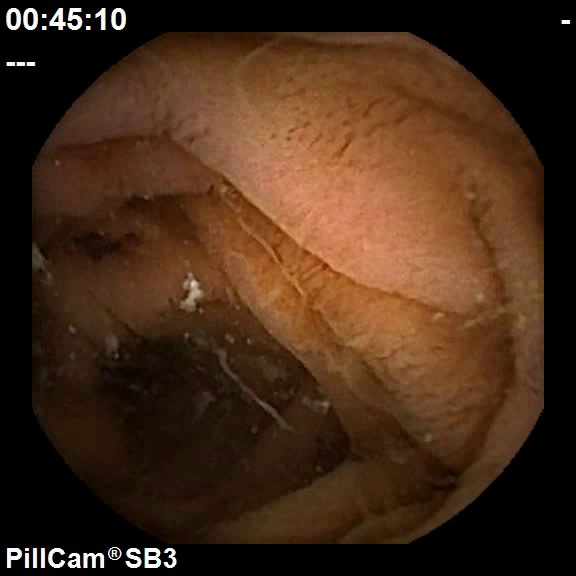}
     \end{subfigure}
    \hspace{-1.35cm}
      \begin{subfigure}[b]{0.1\textwidth}
         \centering
         \includegraphics[trim=32 32 32 32,clip,scale=0.04]{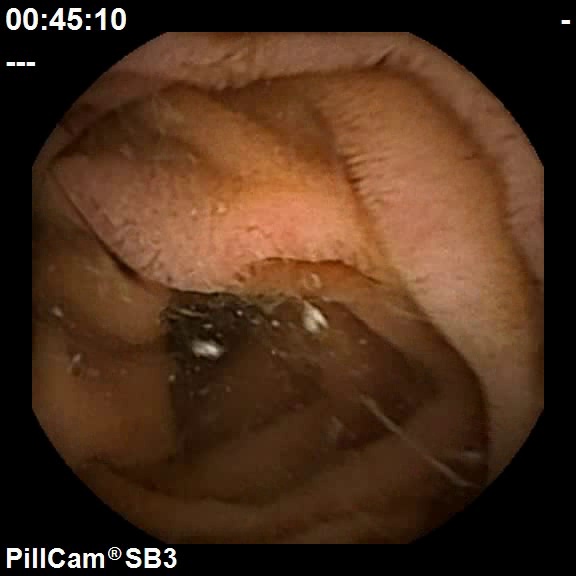}
     \end{subfigure}    
     \hspace{-1.35cm}
      \begin{subfigure}[b]{0.1\textwidth}
         \centering
         \includegraphics[trim=32 32 32 32,clip,scale=0.04]{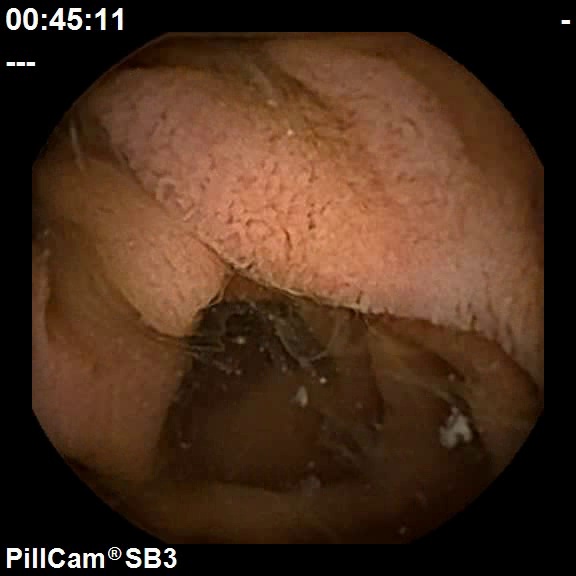}
     \end{subfigure}
    \hspace{-1.35cm}
      \begin{subfigure}[b]{0.1\textwidth}
         \centering
         \includegraphics[trim=32 32 32 32,clip,scale=0.04]{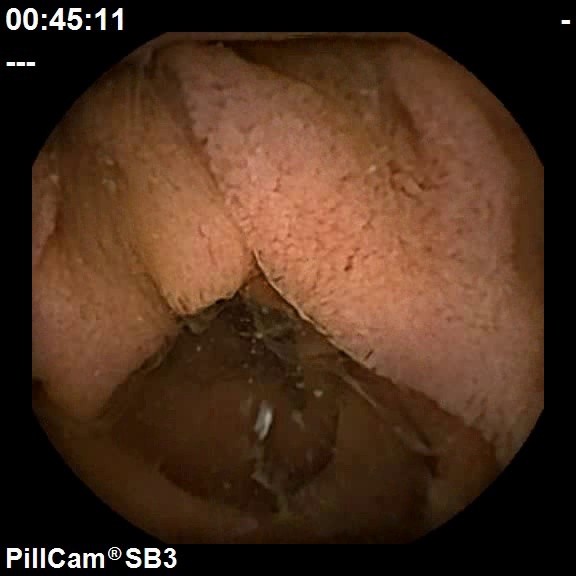}
     \end{subfigure}
     \hspace{-1.35cm}
    \begin{subfigure}[b]{0.1\textwidth}
         \centering
         \includegraphics[trim=32 32 32 32,clip,scale=0.04]{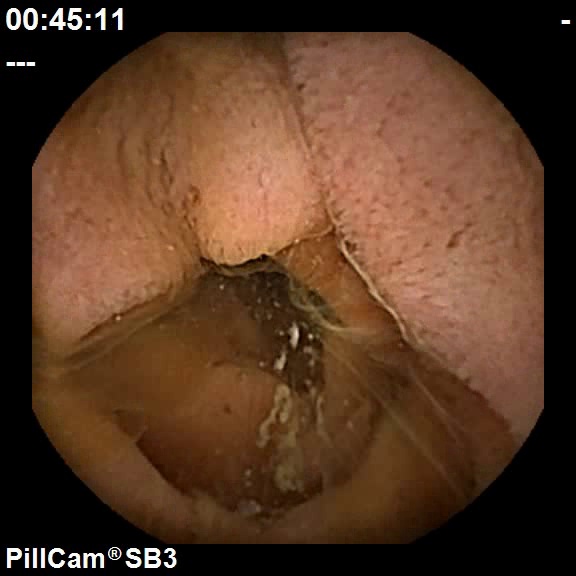}
     \end{subfigure}
     \hspace{-1.35cm}
      \begin{subfigure}[b]{0.1\textwidth}
         \centering
         \includegraphics[trim=32 32 32 32,clip,scale=0.04]{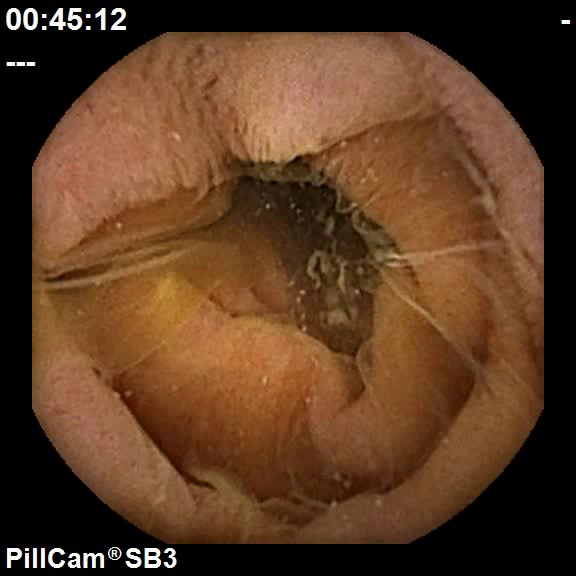}
     \end{subfigure}
     \hspace{-1.35cm}
      \begin{subfigure}[b]{0.1\textwidth}
         \centering
         \includegraphics[trim=32 32 32 32,clip,scale=0.04]{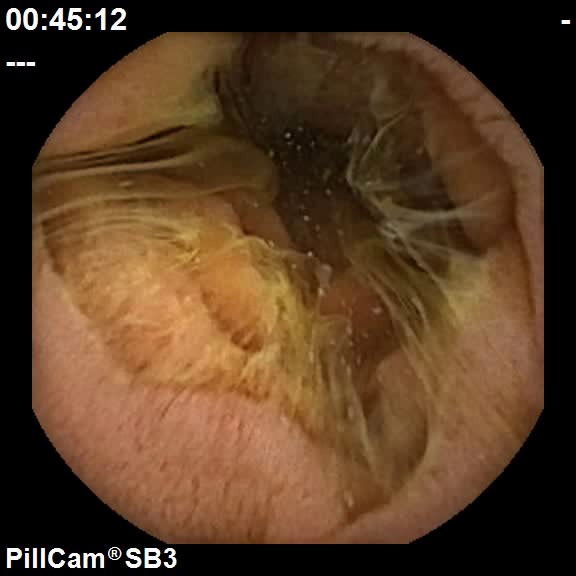}
     \end{subfigure}
    \hspace{-1.35cm}
      \begin{subfigure}[b]{0.1\textwidth}
         \centering
         \includegraphics[trim=32 32 32 32,clip,scale=0.04]{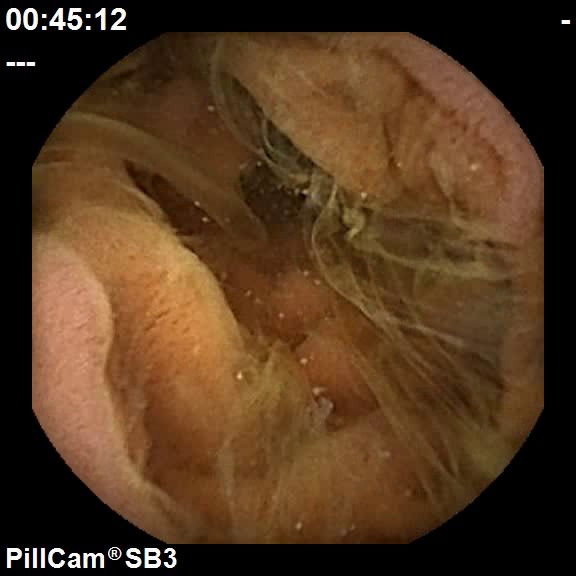}
     \end{subfigure}
    \hspace{-1.35cm}
      \begin{subfigure}[b]{0.1\textwidth}
         \centering
         \includegraphics[trim=32 32 32 32,clip,scale=0.04]{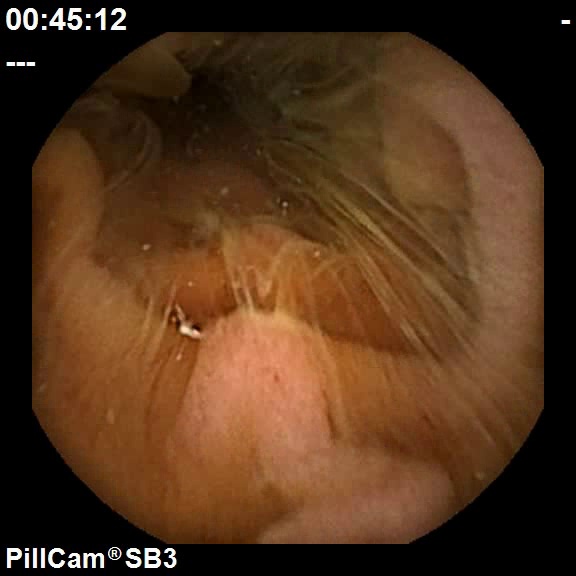}
     \end{subfigure}
    \hspace{-.750cm}
   \rule{1pt}{23pt}
    \hspace{-.750cm}
      \begin{subfigure}[b]{0.1\textwidth}
         \centering
         \includegraphics[trim=32 32 32 32,clip,scale=0.04]{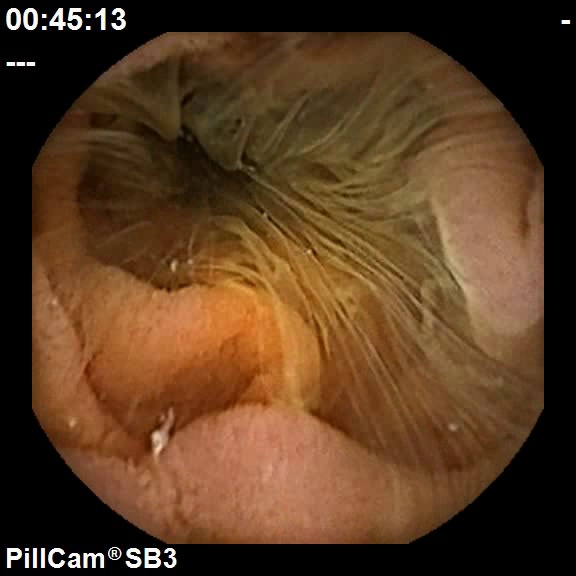}
     \end{subfigure}
    \hspace{-1.35cm}
      \begin{subfigure}[b]{0.1\textwidth}
         \centering
         \includegraphics[trim=32 32 32 32,clip,scale=0.04]{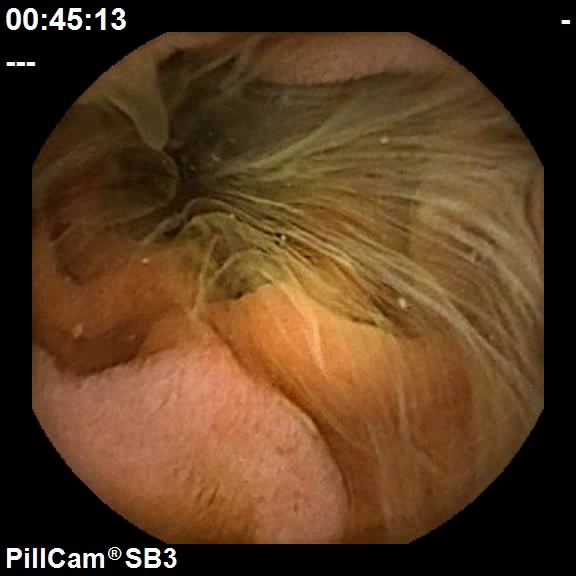}
     \end{subfigure}
    \hspace{-1.35cm}
      \begin{subfigure}[b]{0.1\textwidth}
         \centering
         \includegraphics[trim=32 32 32 32,clip,scale=0.04]{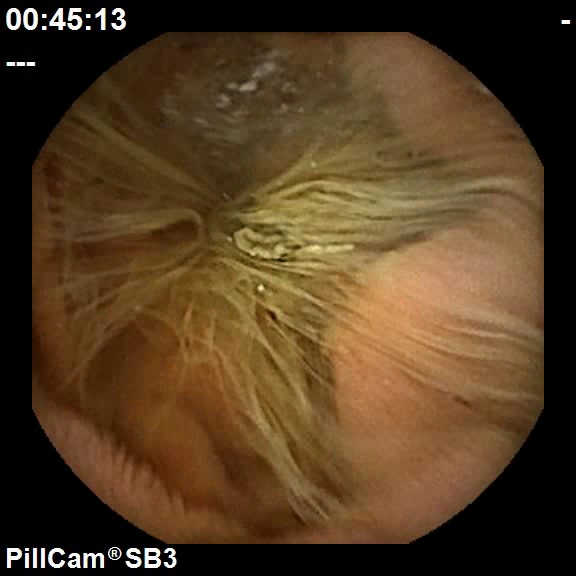}
     \end{subfigure}
    \hspace{-.750cm}
   \rule{1pt}{23pt}
    \hspace{-.750cm}
      \begin{subfigure}[b]{0.1\textwidth}
         \centering
         \includegraphics[trim=32 32 32 32,clip,scale=0.04]{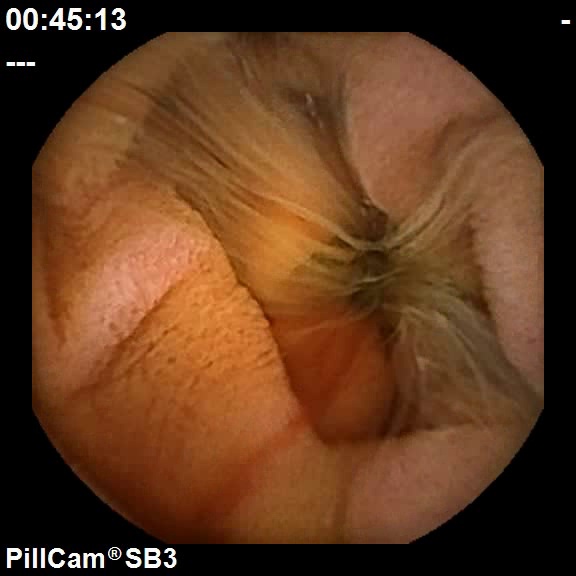}
     \end{subfigure}
     \hspace{-1.35cm}
      \begin{subfigure}[b]{0.1\textwidth}
         \centering
         \includegraphics[trim=32 32 32 32,clip,scale=0.04]{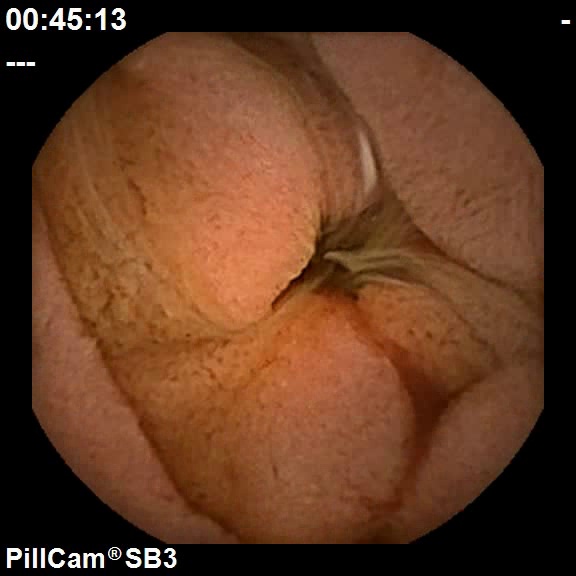}
     \end{subfigure}
     \hspace{-1.35cm}
      \begin{subfigure}[b]{0.1\textwidth}
         \centering
         \includegraphics[trim=32 32 32 32,clip,scale=0.04]{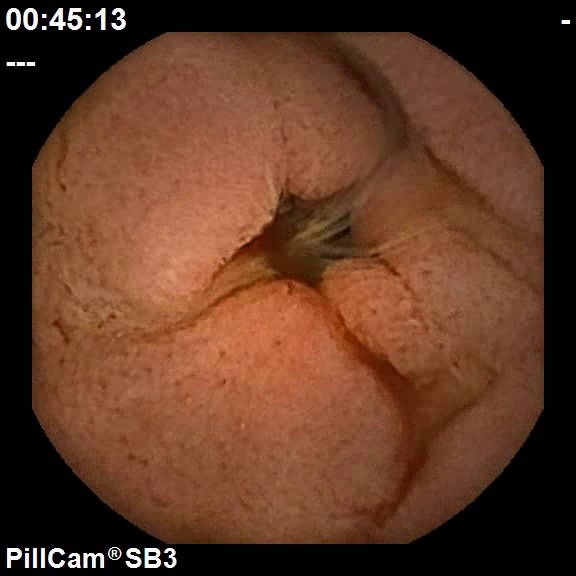}
     \end{subfigure}
     \hspace{-1.35cm}
      \begin{subfigure}[b]{0.1\textwidth}
         \centering
         \includegraphics[trim=32 32 32 32,clip,scale=0.04]{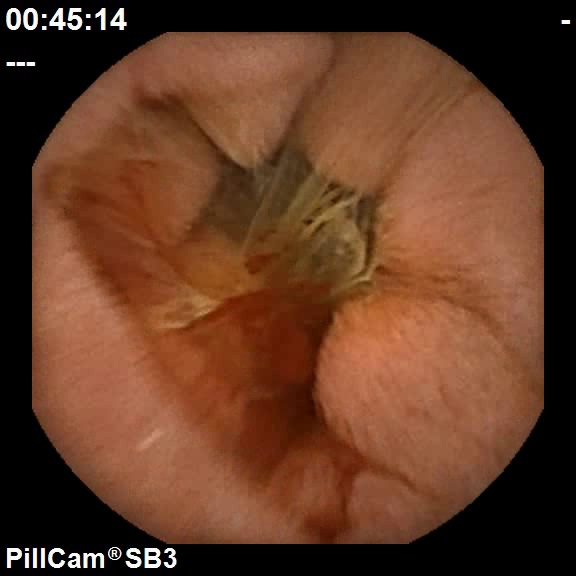}
     \end{subfigure}
    \hspace{-.750cm}
   \rule{1pt}{23pt}
    \hspace{-.750cm}
      \begin{subfigure}[b]{0.1\textwidth}
         \centering
         \includegraphics[trim=32 32 32 32,clip,scale=0.04]{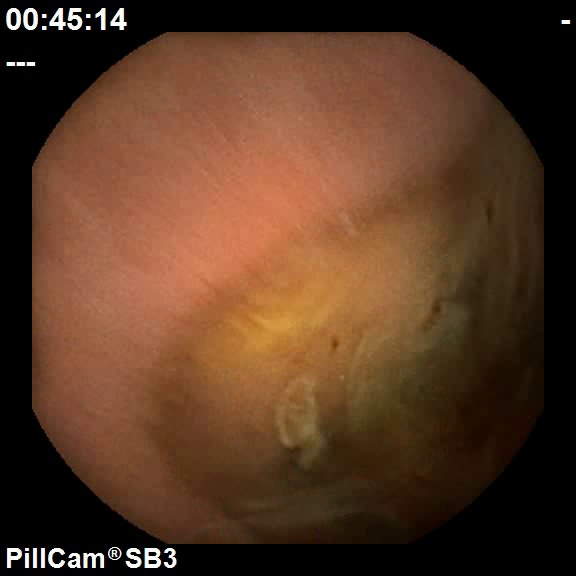}
     \end{subfigure}
    \hspace{-1.35cm}
      \begin{subfigure}[b]{0.1\textwidth}
         \centering
         \includegraphics[trim=32 32 32 32,clip,scale=0.04]{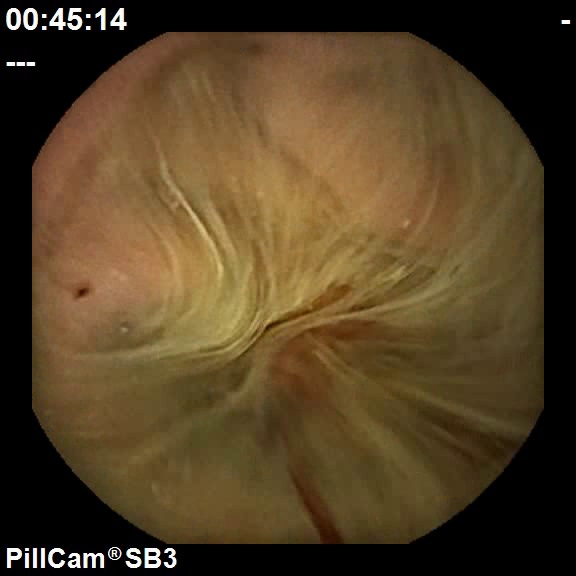}
     \end{subfigure}
    \hspace{-1.35cm}
      \begin{subfigure}[b]{0.1\textwidth}
         \centering
         \includegraphics[trim=32 32 32 32,clip,scale=0.04]{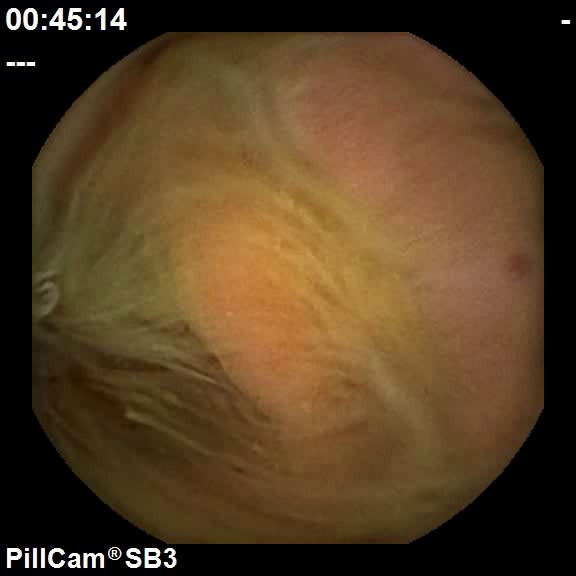}
     \end{subfigure}
    \hspace{-1.35cm}
      \begin{subfigure}[b]{0.1\textwidth}
         \centering
         \includegraphics[trim=32 32 32 32,clip,scale=0.04]{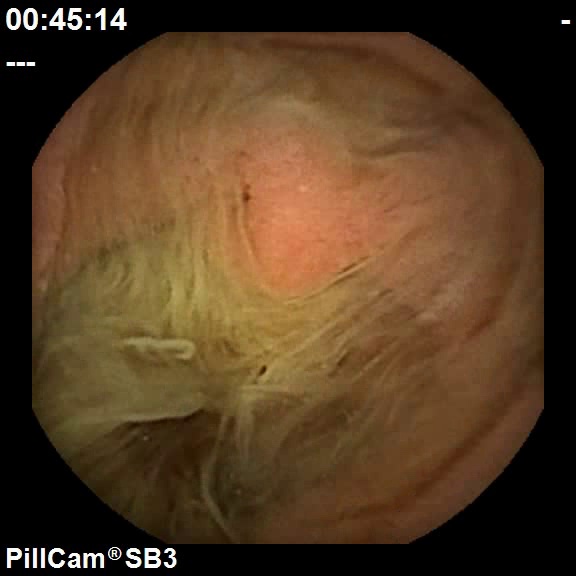}
     \end{subfigure}
    \hspace{-.750cm}
   \rule{1pt}{23pt}
    \hspace{-.750cm}
    \begin{subfigure}[b]{0.1\textwidth}
         \centering
         \includegraphics[trim=32 32 32 32,clip,scale=0.04]{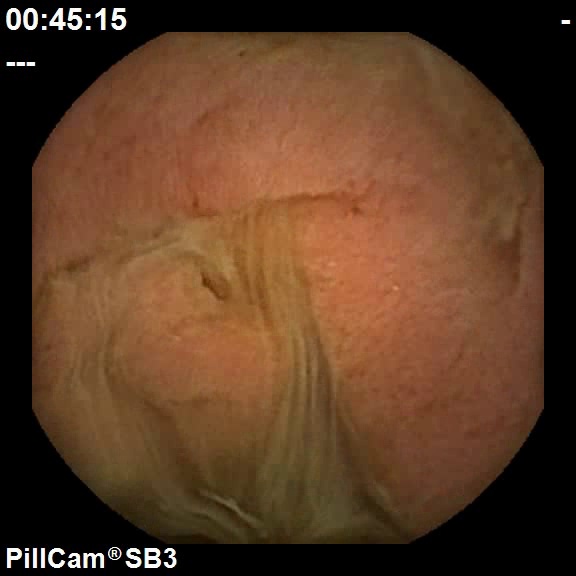}
     \end{subfigure}
     \hspace{-1.35cm}
      \begin{subfigure}[b]{0.1\textwidth}
         \centering
         \includegraphics[trim=32 32 32 32,clip,scale=0.04]{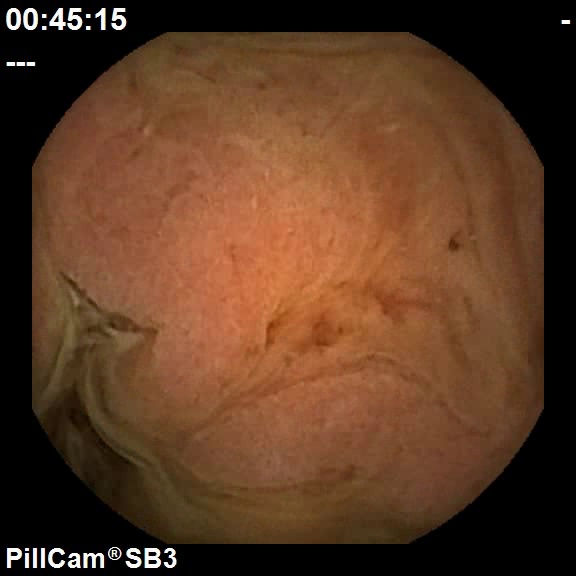}
     \end{subfigure}
     \hspace{-1.35cm}
      \begin{subfigure}[b]{0.1\textwidth}
         \centering
         \includegraphics[trim=32 32 32 32,clip,scale=0.04]{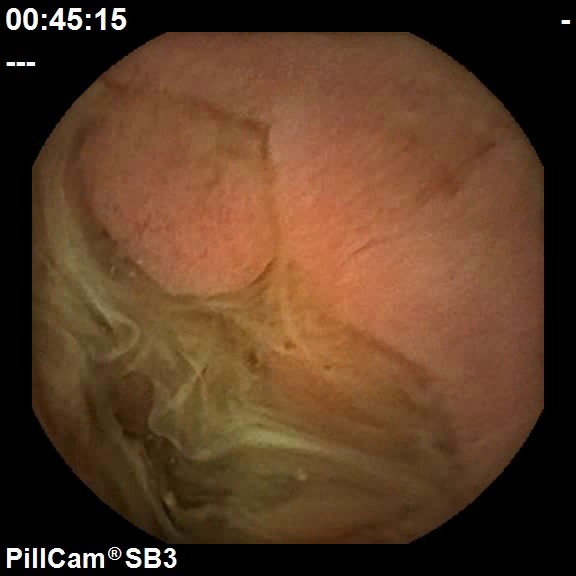}
     \end{subfigure}
    \hspace{-.750cm}
   \rule{1pt}{23pt}
    \hspace{-.750cm}
      \begin{subfigure}[b]{0.1\textwidth}
         \centering
         \includegraphics[trim=32 32 32 32,clip,scale=0.04]{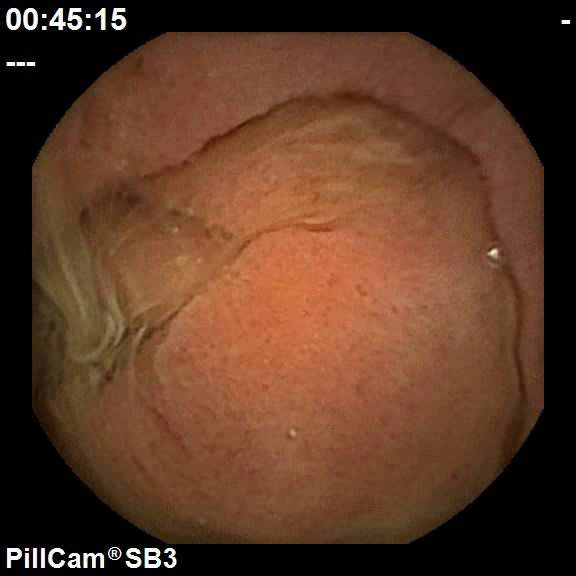}
     \end{subfigure}
    \caption{Visual Illustration of Detected Video Boundaries}
    \label{fig:sbds}
\end{figure}

As shown in figure \ref{fig:sbds}, some of the boundaries detected in the sequence of video frames are not necessarily indicative of pathological change event. However, very similar frames are captured in the same temporal boundaries. Clearly, detecting pathological boundaries in VCE videos is not trivial and also a very challenging problem. Therefore, a binary classification model that can encode the abnormalities into binary category may help mitigate this challenge.

\section*{Conclusion and Future Works}
In this paper, we developed a novel unsupervised technique for temporal segmentation of long capsule endoscopy videos. While our method can be generalized to videos across other domains, we experimented using capsule endoscopy videos collected from patients during real clinical examination. All collected data went through proper IRB approval prior to analysis. After downloading the video from the Rapid reader software, we extracted features from each frame using a pre-trained CNN model. The high-dimensional frame features were projected into lower 1-dimensional representation for the entire video. We applied the pruned exact linear time algorithm to detect transition boundaries in the video using this lower dimensional embedding. Our result showed that the transition detection algorithm is able to better capture pathological event in the sequence of frames when the PCA was used as the embedding mechanism. PCA achieved an AUC-ROC of 66\% and outperformed other non-linear embedding techniques. While our method can easily generalize across multiple domains, when applied on CE videos, the proposed technique can facilitate experts' review of CE videos through significant time and effort saving. For our next step, we will develop a fully integrated long video summarization model requiring little of no expert supervision.

% \section*{Acknowledgements}

% The project is funded by the Army Research Lab under Research Funding Source Award Number: W911NF1820279. We thank the Data Science Institute of the University of Virginia for the support received for this project.

\bibliographystyle{IEEEtran} 
\bibliography{main}

% Generated by IEEEtran.bst, version: 1.14 (2015/08/26)
\begin{thebibliography}{10}
\providecommand{\url}[1]{#1}
\csname url@samestyle\endcsname
\providecommand{\newblock}{\relax}
\providecommand{\bibinfo}[2]{#2}
\providecommand{\BIBentrySTDinterwordspacing}{\spaceskip=0pt\relax}
\providecommand{\BIBentryALTinterwordstretchfactor}{4}
\providecommand{\BIBentryALTinterwordspacing}{\spaceskip=\fontdimen2\font plus
\BIBentryALTinterwordstretchfactor\fontdimen3\font minus
  \fontdimen4\font\relax}
\providecommand{\BIBforeignlanguage}[2]{{%
\expandafter\ifx\csname l@#1\endcsname\relax
\typeout{** WARNING: IEEEtran.bst: No hyphenation pattern has been}%
\typeout{** loaded for the language `#1'. Using the pattern for}%
\typeout{** the default language instead.}%
\else
\language=\csname l@#1\endcsname
\fi
#2}}
\providecommand{\BIBdecl}{\relax}
\BIBdecl

\bibitem{iddan2000wireless}
G.~Iddan, G.~Meron, A.~Glukhovsky, and P.~Swain, ``Wireless capsule
  endoscopy,'' \emph{Nature}, vol. 405, no. 6785, pp. 417--417, 2000.

\bibitem{swain2003wireless}
P.~Swain, ``Wireless capsule endoscopy,'' \emph{Gut}, vol.~52, no. suppl 4, pp.
  iv48--iv50, 2003.

\bibitem{Simonyan2014Sep}
\BIBentryALTinterwordspacing
K.~Simonyan and A.~Zisserman, ``{Very Deep Convolutional Networks for
  Large-Scale Image Recognition},'' \emph{arXiv}, Sep 2014. [Online].
  Available: \url{https://arxiv.org/abs/1409.1556v6}
\BIBentrySTDinterwordspacing

\bibitem{chen2017deep}
M.~Chen, X.~Shi, Y.~Zhang, D.~Wu, and M.~Guizani, ``Deep features learning for
  medical image analysis with convolutional autoencoder neural network,''
  \emph{IEEE Transactions on Big Data}, 2017.

\bibitem{sali2020hierarchical}
R.~Sali, S.~Adewole, L.~Ehsan, L.~A. Denson, P.~Kelly, B.~C. Amadi, L.~Holtz,
  S.~A. Ali, S.~R. Moore, S.~Syed \emph{et~al.}, ``Hierarchical deep
  convolutional neural networks for multi-category diagnosis of
  gastrointestinal disorders on histopathological images,'' \emph{arXiv
  preprint arXiv:2005.03868}, 2020.

\bibitem{malik2021ten}
A.~Malik, P.~Patel, L.~Ehsan, S.~Guleria, T.~Hartka, S.~Adewole, and S.~Syed,
  ``Ten simple rules for engaging with artificial intelligence in
  biomedicine,'' 2021.

\bibitem{gao2020deep}
Y.~Gao, W.~Lu, X.~Si, and Y.~Lan, ``Deep model-based semi-supervised learning
  way for outlier detection in wireless capsule endoscopy images,'' \emph{IEEE
  Access}, vol.~8, pp. 81\,621--81\,632, 2020.

\bibitem{smeaton2010video}
A.~F. Smeaton, P.~Over, and A.~R. Doherty, ``Video shot boundary detection:
  Seven years of trecvid activity,'' \emph{Computer Vision and Image
  Understanding}, vol. 114, no.~4, pp. 411--418, 2010.

\bibitem{geisler2000open}
G.~Geisler and G.~Marchionini, ``The open video project: research-oriented
  digital video repository,'' in \emph{Proceedings of the fifth ACM conference
  on Digital libraries}, 2000, pp. 258--259.

\bibitem{shou2016temporal}
Z.~Shou, D.~Wang, and S.-F. Chang, ``Temporal action localization in untrimmed
  videos via multi-stage cnns,'' in \emph{Proceedings of the IEEE conference on
  computer vision and pattern recognition}, 2016, pp. 1049--1058.

\bibitem{gao2017video}
L.~Gao, Z.~Guo, H.~Zhang, X.~Xu, and H.~T. Shen, ``Video captioning with
  attention-based lstm and semantic consistency,'' \emph{IEEE Transactions on
  Multimedia}, vol.~19, no.~9, pp. 2045--2055, 2017.

\bibitem{rahim2020survey}
T.~Rahim, M.~A. Usman, and S.~Y. Shin, ``A survey on contemporary
  computer-aided tumor, polyp, and ulcer detection methods in wireless capsule
  endoscopy imaging,'' \emph{Computerized Medical Imaging and Graphics}, p.
  101767, 2020.

\bibitem{rasoul2021feature}
S.~Rasoul, S.~Adewole, and A.~Akakpo, ``Feature selection using reinforcement
  learning,'' \emph{arXiv preprint arXiv:2101.09460}, 2021.

\bibitem{adewole2020deep}
S.~Adewole, M.~Yeghyayan, D.~Hyatt, L.~Ehsan, J.~Jablonski, A.~Copland,
  S.~Syed, and D.~Brown, ``Deep learning methods for anatomical landmark
  detection in video capsule endoscopy images,'' in \emph{Proceedings of the
  Future Technologies Conference}.\hskip 1em plus 0.5em minus 0.4em\relax
  Springer, 2020, pp. 426--434.

\bibitem{chen2016wireless}
J.~Chen, Y.~Zou, and Y.~Wang, ``Wireless capsule endoscopy video summarization:
  a learning approach based on siamese neural network and support vector
  machine,'' in \emph{2016 23rd International Conference on Pattern Recognition
  (ICPR)}.\hskip 1em plus 0.5em minus 0.4em\relax IEEE, 2016, pp. 1303--1308.

\bibitem{adewole2021lesion2vec}
S.~Adewole, P.~Fernandez, J.~Jablonski, S.~Syed, A.~Copland, M.~Porter, and
  D.~Brown, ``Lesion2vec: Deep metric learning for few shot multiple lesions
  recognition in wireless capsule endoscopy,'' \emph{arXiv preprint
  arXiv:2101.04240}, 2021.

\bibitem{sainju2014automated}
S.~Sainju, F.~M. Bui, and K.~A. Wahid, ``Automated bleeding detection in
  capsule endoscopy videos using statistical features and region growing,''
  \emph{Journal of medical systems}, vol.~38, no.~4, p.~25, 2014.

\bibitem{mamonov2014automated}
A.~V. Mamonov, I.~N. Figueiredo, P.~N. Figueiredo, and Y.-H.~R. Tsai,
  ``Automated polyp detection in colon capsule endoscopy,'' \emph{IEEE
  transactions on medical imaging}, vol.~33, no.~7, pp. 1488--1502, 2014.

\bibitem{yuan2015saliency}
Y.~Yuan, J.~Wang, B.~Li, and M.~Q.-H. Meng, ``Saliency based ulcer detection
  for wireless capsule endoscopy diagnosis,'' \emph{IEEE transactions on
  medical imaging}, vol.~34, no.~10, pp. 2046--2057, 2015.

\bibitem{tsuboi2020artificial}
A.~Tsuboi, S.~Oka, K.~Aoyama, H.~Saito, T.~Aoki, A.~Yamada, T.~Matsuda,
  M.~Fujishiro, S.~Ishihara, M.~Nakahori \emph{et~al.}, ``Artificial
  intelligence using a convolutional neural network for automatic detection of
  small-bowel angioectasia in capsule endoscopy images,'' \emph{Digestive
  Endoscopy}, vol.~32, no.~3, pp. 382--390, 2020.

\bibitem{pogorelov2018deep}
K.~Pogorelov, O.~Ostroukhova, M.~Jeppsson, H.~Espeland, C.~Griwodz,
  T.~de~Lange, D.~Johansen, M.~Riegler, and P.~Halvorsen, ``Deep learning and
  hand-crafted feature based approaches for polyp detection in medical
  videos,'' in \emph{2018 IEEE 31st International Symposium on Computer-Based
  Medical Systems (CBMS)}.\hskip 1em plus 0.5em minus 0.4em\relax IEEE, 2018,
  pp. 381--386.

\bibitem{zhao2010abnormality}
Q.~Zhao and M.~Q.-H. Meng, ``An abnormality based wce video segmentation
  strategy,'' in \emph{2010 IEEE International Conference on Automation and
  Logistics}.\hskip 1em plus 0.5em minus 0.4em\relax IEEE, 2010, pp. 565--570.

\bibitem{iakovidis2010reduction}
D.~K. Iakovidis, S.~Tsevas, and A.~Polydorou, ``Reduction of capsule endoscopy
  reading times by unsupervised image mining,'' \emph{Computerized Medical
  Imaging and Graphics}, vol.~34, no.~6, pp. 471--478, 2010.

\bibitem{emam2015adaptive}
A.~Z. Emam, Y.~A. Ali, and M.~M.~B. Ismail, ``Adaptive features extraction for
  capsule endoscopy (ce) video summarization,'' in \emph{International
  Conference on Computer Vision and Image Analysis Applications}.\hskip 1em
  plus 0.5em minus 0.4em\relax IEEE, 2015, pp. 1--5.

\bibitem{mehmood2014video}
I.~Mehmood, M.~Sajjad, and S.~W. Baik, ``Video summarization based
  tele-endoscopy: a service to efficiently manage visual data generated during
  wireless capsule endoscopy procedure,'' \emph{Journal of medical systems},
  vol.~38, no.~9, p. 109, 2014.

\bibitem{mohammed2017sparse}
A.~Mohammed, S.~Yildirim, M.~Pedersen, {\O}.~Hovde, and F.~Cheikh, ``Sparse
  coded handcrafted and deep features for colon capsule video summarization,''
  in \emph{2017 IEEE 30th International Symposium on Computer-Based Medical
  Systems (CBMS)}.\hskip 1em plus 0.5em minus 0.4em\relax IEEE, 2017, pp.
  728--733.

\bibitem{ismail2013endoscopy}
M.~M.~B. Ismail, O.~Bchir, and A.~Z. Emam, ``Endoscopy video summarization
  based on unsupervised learning and feature discrimination,'' in \emph{2013
  Visual Communications and Image Processing (VCIP)}.\hskip 1em plus 0.5em
  minus 0.4em\relax IEEE, 2013, pp. 1--6.

\bibitem{vu2009detection}
H.~Vu, T.~Echigo, R.~Sagawa, K.~Yagi, M.~Shiba, K.~Higuchi, T.~Arakawa, and
  Y.~Yagi, ``Detection of contractions in adaptive transit time of the small
  bowel from wireless capsule endoscopy videos,'' \emph{Computers in biology
  and medicine}, vol.~39, no.~1, pp. 16--26, 2009.

\bibitem{mackiewicz2008wireless}
M.~Mackiewicz, J.~Berens, and M.~Fisher, ``Wireless capsule endoscopy color
  video segmentation,'' \emph{IEEE Transactions on Medical Imaging}, vol.~27,
  no.~12, pp. 1769--1781, 2008.

\bibitem{chen2009developing}
Y.-j. Chen, W.~Yasen, J.~Lee, D.~Lee, and Y.~Kim, ``Developing assessment
  system for wireless capsule endoscopy videos based on event detection,'' in
  \emph{Medical Imaging 2009: Computer-Aided Diagnosis}, vol. 7260.\hskip 1em
  plus 0.5em minus 0.4em\relax International Society for Optics and Photonics,
  2009, p. 72601G.

\bibitem{sharma2019data}
V.~Sharma, B.~Shpringer, S.~M. Yang, M.~Bolger, S.~Adewole, D.~Brown, and
  E.~Gharavi, ``Data collection methods for building a free response training
  simulation,'' in \emph{2019 Systems and Information Engineering Design
  Symposium (SIEDS)}.\hskip 1em plus 0.5em minus 0.4em\relax IEEE, 2019, pp.
  1--6.

\bibitem{malladi2013online}
R.~Malladi, G.~P. Kalamangalam, and B.~Aazhang, ``Online bayesian change point
  detection algorithms for segmentation of epileptic activity,'' in \emph{2013
  Asilomar Conference on Signals, Systems and Computers}.\hskip 1em plus 0.5em
  minus 0.4em\relax IEEE, 2013, pp. 1833--1837.

\bibitem{reeves2007review}
J.~Reeves, J.~Chen, X.~L. Wang, R.~Lund, and Q.~Q. Lu, ``A review and
  comparison of changepoint detection techniques for climate data,''
  \emph{Journal of applied meteorology and climatology}, vol.~46, no.~6, pp.
  900--915, 2007.

\bibitem{chowdhury2012bayesian}
M.~F.~R. Chowdhury, S.-A. Selouani, and D.~O’Shaughnessy, ``Bayesian on-line
  spectral change point detection: a soft computing approach for on-line asr,''
  \emph{International Journal of Speech Technology}, vol.~15, no.~1, pp. 5--23,
  2012.

\bibitem{cho2015multiple}
H.~Cho and P.~Fryzlewicz, ``Multiple-change-point detection for high
  dimensional time series via sparsified binary segmentation,'' \emph{Journal
  of the Royal Statistical Society: Series B: Statistical Methodology}, pp.
  475--507, 2015.

\bibitem{aminikhanghahi2017survey}
S.~Aminikhanghahi and D.~J. Cook, ``A survey of methods for time series change
  point detection,'' \emph{Knowledge and information systems}, vol.~51, no.~2,
  pp. 339--367, 2017.

\bibitem{tartakovsky2014sequential}
A.~Tartakovsky, I.~Nikiforov, and M.~Basseville, \emph{Sequential analysis:
  Hypothesis testing and changepoint detection}.\hskip 1em plus 0.5em minus
  0.4em\relax CRC Press, 2014.

\bibitem{Frade-2007-9831}
F.~D. la~Torre~Frade, J.~Campoy, Z.~Ambadar, and J.~F. Cohn, ``Temporal
  segmentation of facial behavior,'' in \emph{Proceedings of (ICCV)
  International Conference on Computer Vision}, October 2007.

\bibitem{Lafferty2001Jun}
J.~D. Lafferty, A.~McCallum, and F.~C.~N. Pereira, ``{Conditional Random
  Fields: Probabilistic Models for Segmenting and Labeling Sequence Data},'' in
  \emph{{ICML '01: Proceedings of the Eighteenth International Conference on
  Machine Learning}}.\hskip 1em plus 0.5em minus 0.4em\relax San Francisco, CA,
  USA: Morgan Kaufmann Publishers Inc., Jun 2001, pp. 282--289.

\bibitem{adewole2020dialogue}
S.~Adewole, E.~Gharavi, B.~Shpringer, M.~Bolger, V.~Sharma, S.~M. Yang, and
  D.~E. Brown, ``Dialogue-based simulation for cultural awareness training,''
  \emph{arXiv preprint arXiv:2002.00223}, 2020.

\bibitem{Chen2012}
J.~Chen and A.~K. Gupta, \emph{{Parametric Statistical Change Point
  Analysis}}.\hskip 1em plus 0.5em minus 0.4em\relax Basel, Switzerland:
  Birkh{\ifmmode\ddot{a}\else\"{a}\fi}user, 2012.

\bibitem{csorgHo198820}
M.~Cs{\"o}rg{\H{o}} and L.~Horv{\'a}th, ``20 nonparametric methods for
  changepoint problems,'' \emph{Handbook of statistics}, vol.~7, pp. 403--425,
  1988.

\bibitem{harchaoui2009kernel}
Z.~Harchaoui, E.~Moulines, and F.~R. Bach, ``Kernel change-point analysis,'' in
  \emph{Advances in neural information processing systems}, 2009, pp. 609--616.

\bibitem{Truong2018Jan}
C.~Truong, L.~Oudre, and N.~Vayatis, ``{Selective review of offline change
  point detection methods},'' \emph{arXiv}, Jan 2018.

\bibitem{Rohrbeck2013}
\BIBentryALTinterwordspacing
C.~Rohrbeck, ``{Detection of changes in variance using binary segmentation and
  optimal partitioning},'' 2013, [Online; accessed 17. Jul. 2021]. [Online].
  Available:
  \url{https://www.semanticscholar.org/paper/Detection-of-changes-in-variance-using-binary-and-Rohrbeck/b132976b756634c02a8bdd77d4c530d02222561a}
\BIBentrySTDinterwordspacing

\bibitem{killick2012optimal}
R.~Killick, P.~Fearnhead, and I.~A. Eckley, ``Optimal detection of changepoints
  with a linear computational cost,'' \emph{Journal of the American Statistical
  Association}, vol. 107, no. 500, pp. 1590--1598, 2012.

\bibitem{auger1989algorithms}
I.~E. Auger and C.~E. Lawrence, ``Algorithms for the optimal identification of
  segment neighborhoods,'' \emph{Bulletin of mathematical biology}, vol.~51,
  no.~1, pp. 39--54, 1989.

\bibitem{jackson2005algorithm}
B.~Jackson, J.~D. Scargle, D.~Barnes, S.~Arabhi, A.~Alt, P.~Gioumousis,
  E.~Gwin, P.~Sangtrakulcharoen, L.~Tan, and T.~T. Tsai, ``An algorithm for
  optimal partitioning of data on an interval,'' \emph{IEEE Signal Processing
  Letters}, vol.~12, no.~2, pp. 105--108, 2005.

\bibitem{wold1987principal}
S.~Wold, K.~Esbensen, and P.~Geladi, ``Principal component analysis,''
  \emph{Chemometrics and intelligent laboratory systems}, vol.~2, no. 1-3, pp.
  37--52, 1987.

\bibitem{tschannen2018recent}
M.~Tschannen, O.~Bachem, and M.~Lucic, ``Recent advances in autoencoder-based
  representation learning,'' \emph{arXiv preprint arXiv:1812.05069}, 2018.

\bibitem{van2008visualizing}
L.~Van~der Maaten and G.~Hinton, ``Visualizing data using t-sne.''
  \emph{Journal of machine learning research}, vol.~9, no.~11, 2008.

\bibitem{iobagiu2008colon}
S.~Iobagiu, L.~Ciobanu, and O.~Pascu, ``Colon capsule endoscopy: a new method
  of investigating the large bowel,'' \emph{Journal of Gastrointestinal and
  Liver Diseases}, vol.~17, no.~3, pp. 347--352, 2008.

\end{thebibliography}

\end{document}